\documentclass[%
 reprint,
 amsmath,amssymb,
 aps, nofootinbib,
 floatfix
]{revtex4-1}

\usepackage{interval}
\usepackage{graphicx}  
\usepackage{multirow}
\usepackage{dcolumn}
\usepackage{amsmath, amsfonts, amssymb, amsthm, setspace, array, relsize, verbatim, booktabs, rotating}
\usepackage{bm}
\usepackage{pifont}
\usepackage[table,xcdraw]{xcolor}
\usepackage{soulutf8} 
\usepackage{rotating}  
\usepackage[hidelinks]{hyperref}
\usepackage{scrextend}      

\usepackage{xcolor}
\usepackage{colortbl}

\usepackage{caption}

\usepackage{silence}
\hypersetup{
    colorlinks=False,    
    linkcolor=blue,     
    citecolor=blue,     
    filecolor=blue,     
    urlcolor=blue       
}
\usepackage{float}
\usepackage{makecell}

\begin{document}

\preprint{APS/123-QED}
\title{Discovering equations from data: symbolic regression in dynamical systems}


\author{Beatriz R. Brum}
    \email{beatrizbrum@usp.br} 
        \affiliation{Institute of Sciences Mathematics and Computation, Universidade de São Paulo-Campus de São Carlos, Caixa Postal 668, 13560-970 São Carlos, São Paulo, Brazil}

\author{Luiza Lober}
    \email{luiza.lober@usp.br}
    \affiliation{Institute of Sciences Mathematics and Computation, Universidade de São Paulo-Campus de São Carlos, Caixa Postal 668, 13560-970 São Carlos, São Paulo, Brazil}

\author{Isolde Previdelli}
    \affiliation{Statistic Department, State University of Maringá, Maringá, Brazil.}

\author{Francisco A. Rodrigues}
    \affiliation{Department of Mathematics Applied and Statistics, Institute of Sciences Mathematics and Computation, Universidade de São Paulo-Campus de São Carlos, Caixa Postal 668, 13560-970 São Carlos, São Paulo, Brazil}

\begin{abstract}

The process of discovering equations from data lies at the heart of physics and in many other areas of research, including mathematical ecology and epidemiology. Recently, machine learning methods known as symbolic regression emerged as a way to automate this task. This study presents an overview of the current literature on symbolic regression, while also comparing the efficiency of five state-of-the-art methods in recovering the governing equations from nine processes, including chaotic dynamics and epidemic models. Benchmark results demonstrate the PySR method as the most suitable for inferring equations, with some estimates being indistinguishable from the original analytical forms. These results highlight the potential of symbolic regression as a robust tool for inferring and modeling real-world phenomena.

{\textbf{Keywords}: Symbolic regression, Dynamic processes, Data driven approaches.}
\end{abstract}

\maketitle

\section{\label{sec_intro} Introduction}

The discovery of equations from observational data is one of the fundamental pillars of the traditional scientific method. From the work of Johannes Kepler, who inferred the laws of planetary motion from meticulous astronomical observations \cite{camps2023discovering} collected by Tycho Brahe~\cite{Gleiser}, to Isaac Newton's theoretical formulations that consolidated classical mechanics, the process of identifying mathematical relationships underlying natural phenomena has historically been characterized by its systematic trial-and-error procedures.

In the last few decades, the advent of Big Data, characterized by the production and availability of an immense volume of complex, mostly nonlinear data in several fields, has renewed the need for methods capable of uncovering physical laws directly from data. Understanding the intrinsic structure of these datasets and deriving symbolic representations that capture system-level behavior has intensified the demand for advanced analytical techniques tailored to large-scale
data.
 
Modern computational techniques have accelerated the development and use of various techniques to expand regression methods and propose new approaches. In this context, symbolic regression (SR) has emerged as a powerful tool to automate the discovery of mathematical expressions from data. Unlike parametric methods, which are limited to adjusting coefficients in predefined equations and require extensive analysis of statistical significance, goodness-of-fit, and model diagnostics, SR jointly searches for both the structure of the equation and its parameters. This search is performed by exploring a vast combinatorial space composed of mathematical operators, variables, and constants, which gives SR greater flexibility in modeling complex relationships. 

Symbolic regression models can draw from multiple machine learning techniques, including genetic algorithms \cite{stephens2016genetic}, sparse regression \cite{brunton2016sparse, brunton2016discovering}, neural networks \cite{udrescu2020ai} Kolmogorov-Arnold representations \cite{liu2024kan, liu2024kan2}, and more recently transformer architectures \cite{DAscoli2023Oct}. Some of these are originally conventional algorithms that make no prior assumptions about the model structure \cite{minnebo2011empowering}, limiting human interaction to the definition of basic parameters to execute the algorithm. In contrast, newer models allow for more substantive interaction between the user and the machine. These unconventional approaches make it possible to encode prior knowledge about the underlying dynamics, thereby guiding and constraining the search for interpretable symbolic expressions.

Beyond presenting a historical overview on the evolution of SR techniques, this study applied symbolic regression to data from different dynamical systems. The objective is twofold: to recover the mathematical equations that govern each system, and to compare the performance of several state-of-the-art SR models. Results show that certain algorithms can reconstruct the structural form of specific systems with high accuracy, and in some cases with expressions indistinguishable from the original equations, suggesting great potential for this set of techniques to be used in real world data in future studies.




\subsection{\label{sec:intro sr models} Symbolic regression}
\begin{figure*}[htbp]
        \includegraphics[width=0.75\textwidth]{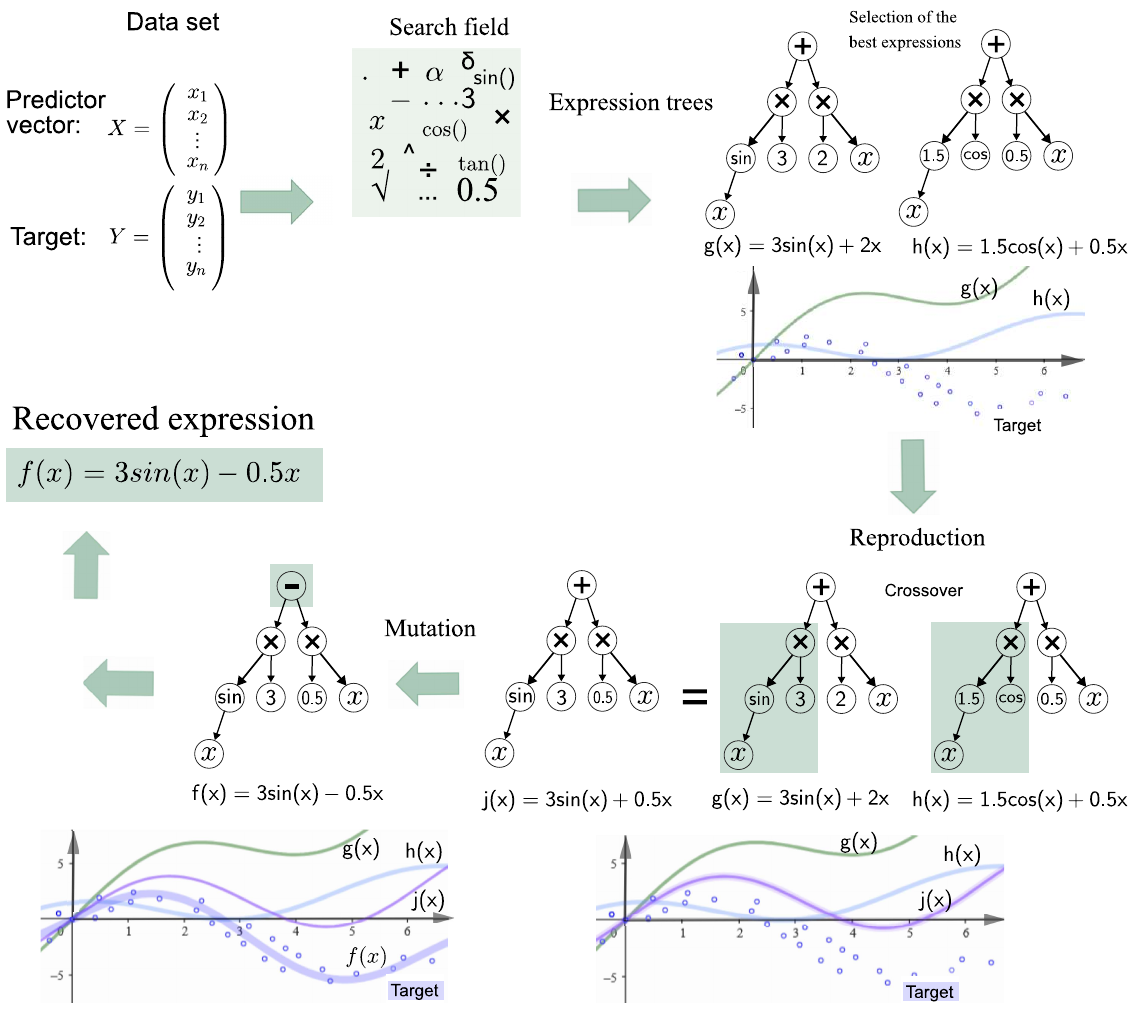}
        \caption{\label{fig:esquema geral} Diagram of the usual process for adjusting a genetic programming-based symbolic regression algorithm. Each method iterates new combinations of functions, tuning model complexity until the best fit is found.} 
\end{figure*}

The first supervised learning method capable of inferring governing equations from data was introduced by \citet{langley1977bacon}, which was known as the BACON algorithm. However, this framework struggled with equations involving constraints, performed poorly with noisy data, and was limited by the modest computing power available in the early 1970s. These factors ultimately led the research community to set aside the idea of automated equation discovery for several decades.

A few years later, in 1975, John Holland and his colleagues at the University of Michigan introduced the concept of genetic algorithms (GAs)~\cite{golberg1989genetic}. Inspired by the principles of natural selection~\cite{darwin1859origin}, GAs provided a robust framework for optimization through evolutionary processes. Building on this idea, \citet{koza1992} introduced genetic programming (GP), which represents candidate solutions as decision trees and iteratively evolves them to discover governing equations for dynamical systems~\cite{diveev2021machine}. These advances created the foundation for symbolic regression, offering a systematic approach to extract mathematical relationships directly from observational data. The usual pipeline of these algorithms is illustrated in Figure \ref{fig:esquema geral}.

In subsequent years, symbolic regression algorithms have gained prominence as they focus on optimization and exploration of approximate solutions for systems with unknown equations of motion \cite{sivanandam2008genetic}. Building on the principles of symbolic regression, dissimilarity analysis, and mating strategies, \citet{gustafson2005} enhanced genetic programming, which yielded demonstrable improvements in accuracy, validated through statistical testing.

A notable milestone in evolutionary SR was introduced by \citet{schmidt2009distilling}, who proposed a method known as Pareto GP, which is a method based on multi-objective optimization that balances accuracy and complexity. In their study, experimental data was collected from computationally tracked motion, which was then used by the model to find the governing equations of that system through either characteristic Lagrangians or Hamiltonians. 


Further progress followed with the development of \textit{Eureqa}, a GP-based algorithm designed to return not only a single expression, but an entire Pareto-optimal set of candidate equations ordered by complexity. Extensive analyses of Eureqa's performance are documented in \citet{dubvcakova2011eureqa}. Continuing in the same evolutionary tradition, \citet{stephens2016genetic} introduced GPlearn, an open-source Python implementation tightly integrated with \texttt{scikit-learn}, facilitating seamless use alongside modern machine learning tools.

Many state-of-the-art symbolic regression algorithms continue to build on GP. Among these, PySR stands out as a powerful tool for generating interpretable models ~\cite{cranmer2023interpretable}. Written in Julia for high-performance computation, PySR is highly optimized while also providing a convenient Python interface, making it both efficient and accessible to researchers.

Parallel to the contributions made by genetic programming, \citet{brunton2016discovering} combined sparsity-promoting techniques and machine learning to discover the governing equations of dynamic systems from noisy data. This methodology allowed their proposed algorithm to be used both in simple and in high-dimensional systems. For instance, it has several applications in both linear and nonlinear oscillatory dynamics, such as chaotic Lorenz systems and fluid vortex shedding behind an obstacle~\cite{brunton2016discovering}. 

Also drawing from sparsity-based algorithms, \citet{mangan2016inferring} developed an algorithm for implicit ordinary differential equations (implicit-SINDy), a contribution that was essential for the discovery of metabolic and regulatory networks, which often exhibit nonlinear dynamics with rational function nonlinearities in their formulation. This algorithm can be applied to biological networks such as Michaelis-Menten enzyme kinetics, a regulatory in the metabolic network for yeast glycolysis. 

In the same group of sparsity-promoting techniques, \citet{rudy2017data} developed a method capable of discovering partial differential equations based on time series measurements, advancing the open field of research of identifiability of dynamical systems.  

Exploring the implementation of sparse regularized regression proposed by \citet{zheng2018unified}, \citet{champion2020unified} presented a sparse optimization framework capable of learning parsimonious models of dynamic systems from data, a formulation that aims to discover the equations of a dynamic system through data and by selecting relevant terms from an array of possible functions. This proposal of a comprehensive framework of symbolic regression models resulted in the ``sparse identification of non linear dynamics" regression package in Python, also known as PySINDy, which also encompasses various optimizers and necessary tools for the operation of the algorithm.
\cite{brunton2016discovering,brunton2016sparse, de2020pysindy,kaptanoglu2021pysindy}. 

Symbolic regression can also benefit from neural network architectures, as first demonstrated by \citet{sahoo2018learning}. Their work demonstrated that neural networks can learn functional forms in mechanical systems such as the cart–pendulum, generalizing beyond the observed parameter space. Later, \citet{udrescu2020ai} introduced AI-Feynman, a multidimensional recursive algorithm combining neural networks with physics-inspired techniques. The approach was validated by effectively discovering all of the 100 selected equations from Feynman's lectures on physics \cite{Feynman_lectures_online}, and as a consequence was considered one of the best algorithms available to investigate physical systems, consequently increasing interest of the scientific community in using SR algorithms in this field.

Among approaches based on neural networks, the PyKAN method~\cite{liu2024kan} offers an interesting alternative to the previously mentioned algorithms. Based on the Kolmogorov–Arnold representation theorem, Kolmogorov–Arnold networks use learnable univariate functions on the edges of the neural network, combined through multivariate linear operations at the nodes, instead of relying on fixed activation functions as in traditional neural networks. This design provides both high expressivity and interpretability, making them well suited for symbolic regression tasks. Extensions such as the MultKAN model~\cite{liu2024kan2} represent some of the most recent advances in SR and highlight the rapid evolution of this area.

Another recently proposed avenue for symbolic regression leverages transformer-based deep neural networks for equation inference, with methods implemented for both functional \cite{Valipour2021Jun, Kamienny2022Nov} and dynamical SR \cite{Kamienny2022Nov, Vastl2022May}. Unlike the aforementioned approaches featuring neural networks, transformers allow for sequence-to-sequence learning of tasks, that is, they translate the trained weights of the model to a different, unseen task, which reduces computational costs in the inference stage.

\subsection{\label{sec:identifiability} Identifiability of systems}
Despite the several approaches to automatic equation discovery that were discussed previously, the question of the overall applicability of symbolic regression, or knowing whether a system is uniquely identifiable from data, remains an active research topic, and an essential consideration when applying these methods to unseen, real world data. 

For parametric ordinary differential equations (ODEs), as considered in this study, one can answer this question using the framework developed by \citet{Qiu2022Oct}, provided that the parameters of those ODEs can assumed linear. Their method also includes quantitative scores that assess the impact of noise in data. The authors also discuss the identifiability for higher-dimensional systems, with the overall conclusions pointing to such systems being identifiable unless their dimensions approach infinity.

In contrast, answering the same question for partial differential equations remains largely unresolved. \citet{Scholl2023} provides a first detailed analysis on this topic, and a few examples of applications are given by \citet{rudy2017data} and \citet{kiyani2023framework}, but the field is still at an early stage. 

More broadly, another relevant issue to equation discovery comes from the need of some systems to apply coordinate transformations to the data to then allow for the equations to be inferred, as was the case for Kepler's laws of planetary motion. Tackling this issue unfortunately requires in-domain knowledge of the system in some cases, although a few methods that will be discussed in the next section do have built-in mechanisms to either simplify equations according to known properties of a system, or use a similar procedure to equation discovery that is usually employed in the domain of physics, which includes transformations to the variables.

\section{\label{subsec_B_SR_description} Symbolic regression algorithms}


The symbolic regression models employed in this study are open source algorithms that represent the most recent and significant contributions to the field. These include: (i) GPlearn, an evolutionary algorithm~\cite{stephens2016genetic}, (ii) AI-Feynman, which combines neural networks with algebraic simplification techniques~\cite{kaptanoglu2021pysindy,udrescu2020ai,udrescu2020ai}, (iii) PySINDy, based on sparse regression~\cite{de2020pysindy}, (iv) PySR, which leverages genetic programming~\cite{cranmer2023interpretable}, (v) PyKAN, a neural-network-based approach~\cite{liu2024kan} and (vi) ODEFormer, applying an encoder-decoder transformer algorithm to obtain dynamical systems equations. The following subsections provide a detailed description of each method

\subsection{\label{sec:GPLearn} GPLearn}

This genetic programming algorithm starts its search by randomly generating a population of symbolic expressions, which are the result of combinations from the search space, or in other words the combination of operators and functions defined by the user, such as $\left\{+, -, *, \sin, \cdots \right\}$. Each of those combinations are treated by the algorithm as a single genetic code, represented by a uniquely coded chromosome.

The evaluation of each individual fit, $f(\boldsymbol{x})$, is then performed through transformations, i.e., genotype-phenotype mappings ($f_g$) and phenotype-fitness ($f_p$) \cite{rothlauf2006representations}, resulting in the convolution $f = f_p(f_g(\boldsymbol{x}^g))$. The fittest individuals, according to a goodness of fit metric selected by the user, are then selected to keep evolving while the others are eliminated \cite{sivanandam2008genetic}. With a new and smaller population, these individuals undergo a selection process in pairs. In the reproduction stage, the algorithm performs crossovers and mutations to generate a new population. This cycle is repeated until convergence, when an expression that balances accuracy and complexity is obtained.

\subsection{\label{sec:AI-Feynman} AI-Feynman} 
The AI-Feynman algorithm, developed at MIT’s Artificial Intelligence Laboratory, is one of the most widely know symbolic regression methods due to its strong descriptive power in physics, as discussed in \autoref{sec:intro sr models}. A key strength of this algorithm lies in the similarity between its procedure and the way physicists traditionally model phenomena. The main advantage it was is the capability of deriving  compositional expressions -- that is, a function $f$ that can be represented as a combination of a small set of elementary functions, e.g., through linear combinations or other structured compositions. 

The set of tools employed by this method ranges from simple processes in the search for symbolic representations, to the application of state-of-the-art machine learning techniques such as neural networks, with the resulting expression being generated through a tree encoding using reverse Polish notation \cite{udrescu2020ai}. The algorithm leverages methods that exploit common simplifying properties in natural physical processes, such as dimensional analysis, a straightforward approach resulting in a considerable reduction of the initial variable space, solving the issue of overshooting the complexity necessary for the expressions. 

Additionally, it also includes nonlinear least squares polynomial fitting, allowing for the adjustment of polynomials ranging from zero to the fourth degree. The method also uses an exhaustive search algorithm, generating a space that includes all possible expressions, from the simplest to the most complex with varied parameter combinations. Moreover, fitting can also be done using neural network-based structures, also simplifying the investigation of symmetry and separability properties. Finally, the algorithm incorporates equality verification and data transformations, returning a list of expressions ordered by the lowest MDL Loss, a criterion based on information theory that selects the shortest hypothesis that strikes a balance between accuracy and complexity. The final choice between them is up to the researcher.  

\subsection{\label{sec:PySINDy} PySINDy} 

This algorithm adds a sparsity constraint in its formulation to identify nonlinear differential equations. This technique also penalizes the error function and facilitates the identification of coefficients with nonlinear behavior by weighting terms according to a set of criteria. The structural form of the aforementioned sparse formulation can be expressed as 
    \begin{eqnarray}
        \frac{d}{dt}x(t) = f(\mathbf{x}(t)).
    \end{eqnarray}
    
The data describing $f$ is required to be sparse in the state variables $\mathbf{x}(t) \in \mathbb{R}^n$ \cite{quade2018sparse}, while the derivative of each of the different variables tends to be sparse in the space of possible functions. For example, in the linear combination below,
    \begin{eqnarray}
        f_i(x) = \xi_1 \theta_1(x)+ \xi_2 \theta_2(x)+ \cdots + \xi_j \theta_j(x),
    \end{eqnarray}
    \noindent
most of the coefficients will end up canceling each other out when an appropriate set of $\theta_j$ functions are found. 

The method then assumes an approximation of the form
    \begin{eqnarray}
        \label{dotX}
        \dot{\mathbf{X}} \approx \Theta(\mathbf{X}) \mathbf{\Xi},
    \end{eqnarray}
    \noindent
where $\mathbf{\Xi}$ is a set of coefficients that determines the active terms in $f$, which must satisfy the sparsity requirement in equation \eqref{dotX} \cite{de2020pysindy,kaptanoglu2021pysindy}.

Thus, the algorithm initializes with the following matrices: \textbf{X} and $\dot{\textbf{X}}$. The first one is composed of time series of the variables measured from the system. The second corresponds to the target matrix of time derivatives, generated by differentiating $\mathbf{X}$ through a set of differentiation methods available in the algorithm.

Finally, the matrix $\Theta(\textbf{X})$ lists the candidate functions for the formulation of the sought symbolic expression, which are user-defined. Through the regularization term $R(\mathbf{\Xi})$, used by the algorithm internally, restrictions are applied to the coefficients of the model's parameters, which characterizes the sparse regression. The parameter $\lambda$ sets the sparsity threshold, and to achieve this, the software also uses the Pareto curve algorithm, which displays the simplest expression identified through regularization, thereby ensuring a more parsimonious model and reducing overfitting.

\subsection{\label{sec:PySR} PySR} 
Based on genetic algorithms and implemented in the Julia programming language, PySR's main advantage over alternatives such as GPLearn lies in the use of a framework with higher computational performance offered by this new programming language, while also standing out for its efficiency and robustness, as discussed in \cite{cranmer2023interpretable}.

Furthermore, it is possible to implement custom functions in the code using SIMD kernels in Julia. These kernels can be applied at runtime, ensuring high performance by processing multiple operations on data simultaneously, performing automatic differentiation, and handling populations of mathematical expressions, while taking advantage of parallel computing \cite{cranmer2023interpretable}. In the aforementioned paper, a benchmarking tool named ``EmpiricalBench" was also proposed, with the objective of comparing, evaluate and measure the capabilities of different symbolic regression algorithms in scientific applications.
    
Another interesting aspect of the algorithm is that simulations with nested operators can be controlled, ensuring that undesired compositions between functions do not occur, which is a useful tool to reduce expression complexity. The method also allows unary operators to be incorporated based on prior theoretical knowledge about the data, in addition to identifying the simplest expression through a combination of precision and complexity, similar to other algorithms. Furthermore, the method returns a list of the best models found, which can be inspected by the researcher if the automatically selected expression is considered inappropriate when properties of the dynamics studied are known.

\subsection{\label{sec:PyKAN} PyKAN} 

PyKAN was developed by \citeauthor{liu2024kan}~\cite{liu2024kan}, replacing multilayer perceptron (MLP) neural networks with Kolmogorov–Arnold networks in its architecture. The motivation is that MLPs are based on the universal approximation theorem, which states that continuous functions can be approximated with arbitrary precision provided that the network has an appropriate topology for that specific dataset. Achieving such a topology, however, is known to be a challenging task, and that often results in highly specialized models that do not generalize well to different problems.

In contrast, the Kolmogorov-Arnold representation theorem guarantees that every continuous function can be expressed as a combination of simpler functions. Given a set of variables derived from a physical process as a continuous function, $f: [0, 1]^n \to \mathbb{R}$, it can be represented as:

    \begin{eqnarray}
        f(x_1, x_2, \ldots, x_n) = \sum_{q=1}^{2n+1} \Phi_q \left( \sum_{p=1}^{n} \phi_{q,p}(x_p) \right),
    \end{eqnarray}
    \noindent
where \(\Phi_q : [0, 1] \to \mathbb{R}\) e \(\phi_{q,p}: \mathbb{R} \to \mathbb{R}\) are continuous functions. This theorem provides a theoretical foundation to Kolmogorov–Arnold networks (KANs) \cite{liu2024kan}. 

In PyKAN's implementation, the authors generalize the original representation, where the depth was 2 layers and the width was equal to $2n + 1$ with $n$ representing the number of input variables. By enabling the construction of deeper architectures and training them via backpropagation, PyKAN can capture more complex relationships commonly encountered in real-world applications \cite{liu2024kan}.

KANs also differs from multi-layer perceptrons (MLPs) by not using fixed activation functions. Instead of conventional weights, KANs adopt one-dimensional functions along the edges, which are parameterized by splines during training. These splines are smooth functions that fit the data, allowing KANs to learn their own activation functions and providing greater flexibility in modeling.

KANs also tends to avoid overfitting: when the network is fed with a training dataset, it learns a specific mapping. However, by using another part of the same dataset, the network learns a different mapping. In practice, the network adjusts only a few control points, ensuring that previously learned information is retained. This is made possible through the control points of B-splines, which allow for local adaptation without compromising the knowledge already acquired, an often observed problem with MLPs. 
During training, the spline grids can be refined, turning them denser and increasing the number of parameters, and as such allowing for more specialized models, allowing the network to specialize while still adjusting primarily a limited number of nodes.

The predominance of the multiplication operation in real physical systems motivated the authors to enhance the algorithm, leading to MultKAN's release. As highlighted by \citet{liu2024kan2}, this new approach has the potential to reveal multiplicative structures present in data. In addition, the authors equipped the algorithm with tools that incorporate prior knowledge about the system under study, such as KANCompiler, providing a significant advantage in discovering symbolic expressions. The method selects the result by minimizing a total cost function that combines the training or test error (MSE) with regularization terms.

\subsection{\label{sec:OdeFormer} ODEFormer} 

To infer dynamical systems using deep learning, ODEFormer, proposed by \citet{DAscoli2023Oct}, is the most up-to-date encoder-decoder transformer method and is supported by an open codebase. Compared with early transformer-based algorithms, ODEFormer has the advantage of being able to infer the governing equations of multidimensional systems, in this way proving to be a particularly well-suited method three dimensional systems and others with several independent variables. Unlike the aforementioned SR algorithms, ODEFormer operates under a different paradigm: its predictive capability is based on dynamics learned during training on large datasets, allowing the algorithm to probabilistically identify the behavior of several systems, later transferred to the dataset of interest.

ODEFormer includes 16 attention heads \cite{Vaswani2017Jun} and 512 embedding dimensions, totalling approximately 86 million parameters. These parameters were trained using an extensive synthetic dataset from various dynamic systems. This data undergo an embedding strategy in which expressions are tokenized and represented as vectors in a space, denoted by $\mathbb{R} ^{((D+1)\times 3)\times d_{emb}}$, where $D$ is the dimension of the original system that generated the data \footnote{$D_{max}=6$ for the pre-trained model used in this study}, and $d_{emb}$ is the embedding dimension associated with the original equation. The ability to treat data as a sequence of tokens, a fundamental principle of modern machine translation, is what makes transformer-based algorithms unique compared to other symbolic regression approaches. To decode and infer equations, the model uses a beam sampling method \cite{VanGael2008Jul}.

The authors also propose an extensive dataset for benchmarking their algorithm, named ODEBench \cite{DAscoli2023Oct}. In it, 63 equations with varying number of dimensions, with some exhibiting chaotic behavior, are used to ascertain the performance of the trained model and released publicly for future benchmarks.



\section{\label{subsec_B_modern_applications} Applications}

Symbolic regression  has been successfully applied to multiple domains of knowledge over the past decade, providing an overall positive contribution to diverse areas of knowledge. Notable examples include climate system modeling \cite{stanislawska2012modeling}, as well the development of hybrid algorithms in materials science \cite{wang2019symbolic}. In ecology, \citet{chen2019revealing} employed genetic programming to understand the dynamics of complex ecosystems, while \citet{abdellaoui2021symbolic} used SR modeling in wind speed prediction. In finance, \citet{luo2023application} used SR with the objective of understanding the implied volatility surface in the financial market.     

Further applications highlight the versatility of SR methods. \citet{kiyani2023framework} applied GPlearn to discover the closed-form of unknown components of complex nonlinear PDEs through a domain decomposition approach. \citet{gudetti2023} demonstrated that SINDy can be used for NVH (noise, vibration and harshness) applications aimed at vibration control in products. \citet{miyazaki2023} employed AI-Feynman to discover the hyperbolic discounting formulation that could not be solved analytically previously. In astrophysics, \citet{wong2022automated} demonstrated that SR through PySR can be used as an effective strategy to identify interpretable gravitational wave population models. Beyond the essence of these applications, there is still other areas that can benefit from this approach, such as epidemic modeling, acoustics and many others. 

The effectiveness of symbolic regression, however, often depends on the nature and structure of the data, which may favor certain algorithms over others. Although highly impactful works have presented increasingly accurate methods for regression, there are still no single standardized methodology to evaluate their efficiency, due to the diversity and complexity in which each symbolic regression model was developed. Some proposed solutions to this problem are discussed below.

\citet{udrescu2020ai}, performed a comprehensive comparison between AI-Feynman and Eureqa \cite{schmidt2014eureqa}, using 120 synthetic datasets for this comparative analysis. More recently, \citet{la2021contemporary} introduced a reproducible and open-source benchmarking project platform named SRBench, which currently evaluates the performance of 14 contemporary symbolic regression methods on 252 datasets, alongside seven machine learning baselines. However, many newer SR algorithms are still missing from this platform at the time of writing.

To better assess the capabilities of newer symbolic regression models, particularly in recovering governing equations and in analyzing state transitions in compartmental epidemiological dynamics, a systematic benchmark was conducted. The epidemiological systems selected for this study, together with other representative dynamical systems of interest, are described in the following section.


\subsection{\label{sec:systems for benchmark} Dynamical systems}
 
In this section, four types of dynamic systems will be discussed: two from the domain of physics, one from biology, and the fourth category consisting of another six systems that describe epidemic propagation through different assumptions on its composing compartments.

The systems presented in Sections \ref{sec:Lorenz} to \ref{sec:lotka-volterra intro} were chosen for their overarching relevance in the domains of physics and biology, allowing this study to analyze the performance of the symbolic regression algorithms in describing a chaotic system, oscillatory motion and predator-prey dynamics, this way anchoring the novel use of symbolic regression algorithms to epidemic models being proposed here to the overall performance of those methods in multiple fields.

\subsubsection{\label{sec:Lorenz} Lorenz attractor}

A classic example of chaotic dynamics that originates from a model to atmospheric convection \cite{Lorenz1963Mar}, and independently from single-mode laser dynamics \cite{Haken1963Jun}, is given by the Lorenz-Haken equations, which are described by

\begin{equation}
    \begin{aligned}
        \dot{x} &=& \sigma (y-x) \\
        \dot{y} &=& -x (\rho - z) -y, \\
        \dot{z} &=& xy - \beta z
    \end{aligned}
\end{equation}
\noindent
where, considering Lorenz's derivation for the atmosphere, $x$ is proportional to the rate of convection of a fluid, $y$ the horizontal temperature variation and $z$ is the vertical temperature variation. $\sigma, \rho$ and $\beta$ are constants that relate to properties of that physical system. The system also appears in many other fields and have generalizations to higher dimensions \cite{Shen2025Jan}. 

\subsubsection{The non-linear pendulum}

\begin{figure}[htbp]
    \centering
    \includegraphics[width=0.5\linewidth]{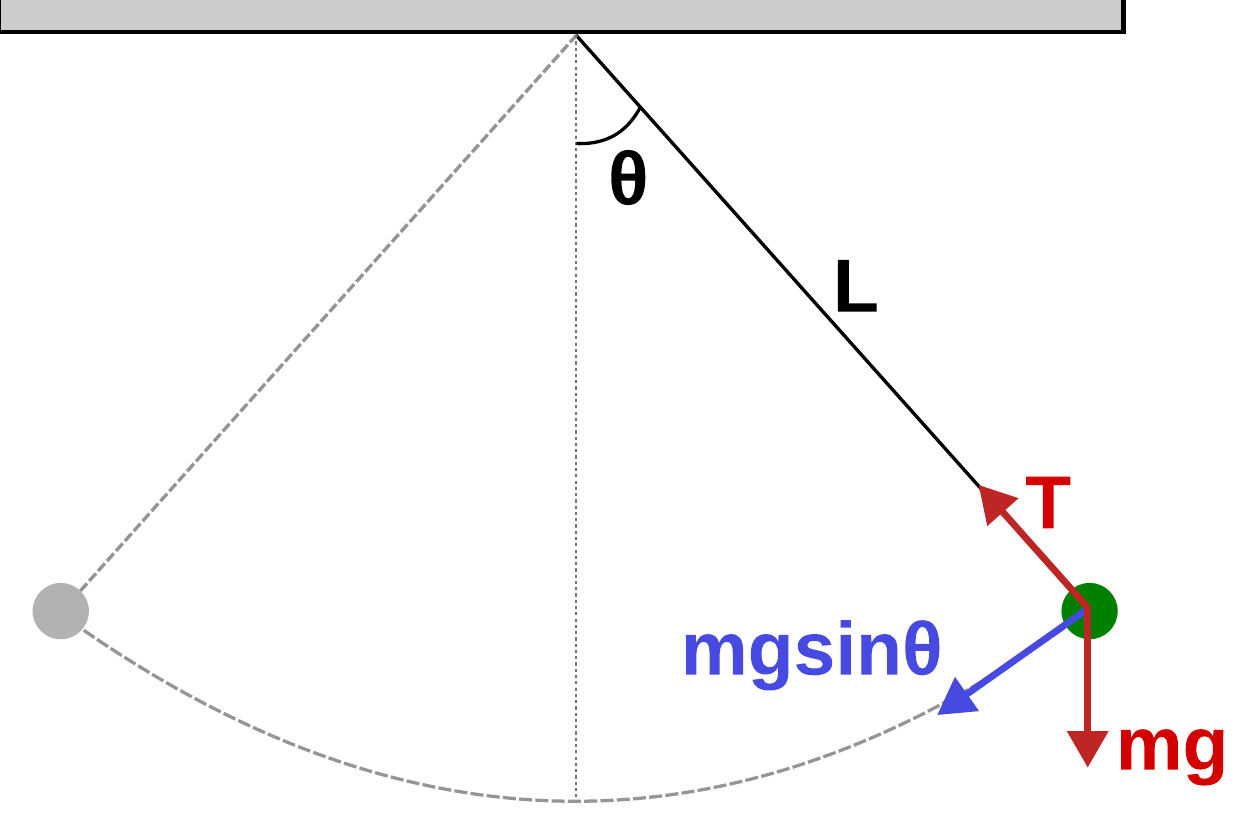}
    \caption{Force diagram of the unitary non-linear pendulum.}
    \label{fig:pendulum diagram}
\end{figure}

Acquiring the equations of motion of this system analytically is relatively simple. To start, drawing the force diagram as done in \autoref{fig:pendulum diagram} already implies that the sum of forces will be $F = mg\text{sin}(\theta)$.  Using Newton's second law of motion and taking $x = r\theta = L\theta$, the previous equation then becomes $F = m \ddot{x} = mL \ddot{\theta} = - \text{mg} \sin (\theta)$. 

Noting that the angular frequency in this case is $\omega = \sqrt{g/L}$ and $\ddot{\theta} + F(\theta) = \ddot{\theta} + \omega^2 \text{sin}\theta = 0$, then by rewriting these terms, one arrives at the following equation:

\begin{equation}
    \begin{aligned}
        \ddot{\theta} = -\frac{g}{l} \text{sin}\theta,
    \end{aligned}   
\end{equation}
\noindent
which, together with $u = \dot{\theta}$, can then be used to simulate the behavior of this system. 

\subsubsection{\label{sec:lotka-volterra intro} Lotka-Volterra predator-prey dynamics}

An essential baseline model in ecology, the Lotka-Volterra equations are used to simulate and understand the interaction of predator and prey populations in a food chain \cite{Diz-Pita2021Jul}. 

Even when making simplistic assumptions on the environment and the populations of the two interacting species, such as ample food being available at all times and that being entirely reliant on the prey population size, no account of genetic variation and adaptation, alongside of no environment changes, the model still is able to capture the general oscillations of population sizes that are observed in nature.

The system of equations that describe this model is given by

\begin{equation}
    \begin{aligned}
        \dot{u} &=& \alpha u - \beta uv \\
        \dot{v} &=& -\gamma v + \delta uv,
    \end{aligned}
\end{equation}
\noindent
where $u$ represents the population density of prey, $v$ the predator's. $\alpha, \beta, \gamma, \delta \in \mathbb{R}^+$, with $\alpha$ as the maximum growth rate of prey per capita, $\beta$ the effect of predators in the death rate of prey and, complementary to the last two, $\gamma$ is the predator's per capita death rate and $\delta$ the effect of prey in predator's growth rate.

When considering the equilibrium of the two populations, that is, $u=\gamma/\delta$ and $y=\alpha/\beta$ respectively for the prey and predator \cite{Lotka-Volterra-Original-Pub}, both depend on each other's parameters, which as a consequence means that increasing the prey growth rate benefits the predator, with that in turn not granting an improvement to the prey population in the long term.


\subsubsection{\label{sec:epi description} Epidemiological compartmental models}

    \begin{table*}
        \centering
        \caption{\label{tab:compartmental models} All compartmental models used and their governing equations, with transition rates as indicated in \autoref{fig:compartmental models schematic}.}
        \resizebox{0.9\linewidth}{!}{
        \begin{tabular}{c|c|c|c|c|c}
        \hline \hline
            SIS & SIR & SIRV & SIRS & SEIR & SEIRD \\ \hline
            
        \begin{tabular}{l}
            $\dot{I} = -\dot{S} = \frac{\beta}{N} IS - \gamma I $ \\
        \end{tabular}   & 
        \begin{tabular}{l}  
            $\dot{S} = -\frac{\beta I S}{N} $ \\ 
            $\dot{I} = \frac{\beta I S}{N} - \gamma I$\\
            $\dot{R} = \gamma I$ \\
        \end{tabular}  & 
        \begin{tabular}{l}
            $\dot{S}= -\frac{\beta S I}{N} - \epsilon S$ \\
            $\dot{I}= \frac{\beta S I}{N} - \gamma I$ \\
            $\dot{R}= \gamma I$ \\
            $\dot{V}= \epsilon S$ \\ 
        \end{tabular}    & 
        \begin{tabular}{l}
            $\dot{S} = -\frac{\beta S I}{N} + \delta R$ \\ 
            $\dot{I} = \frac{\beta S I}{N} - \gamma I$\\
            $\dot{R} = \gamma I - \delta R$ \\ 
        \end{tabular}   &
        \begin{tabular}{l} 
            $\dot{S}=  -\frac{\beta I S}{N} $ \\
            $\dot{E}= \frac{\beta I S}{N} - \sigma  E$ \\
            $\dot{I}= \sigma E - \gamma I$ \\
            $\dot{R}= \gamma I$ \\ 
        \end{tabular}    & 
        \begin{tabular}{l} 
            $\dot{S}= - \frac{\beta S I}{N}$ \\
            $\dot{E}= \frac{\beta S I}{N} - \sigma E$ \\
            $\dot{I}= \sigma E - (\gamma +\mu) I$ \\
            $\dot{R}= \gamma I$ \\
            $\dot{D}= \mu I$ \\ 
        \end{tabular} \\ \hline \hline
        \end{tabular} 
    }
    
\end{table*}

When studying the propagation of diseases in a population, a first step to understand the overall evolution of patients in a group is to compartmentalize each stage of the infection. In doing so, one can not only categorize populations, but also describe rates of transition from each category. 

The simplest model that can be built that way considers only two states: susceptible (S) and infected (I) individuals. Assuming the transition rates of infection ($\beta$), and recovery ($\gamma$), alongside $N$ as the number of individuals, the resulting dynamics will be described by

    \begin{equation}
        \dot{I} = -\dot{S} = \frac{\beta}{N} IS - \gamma I,
    \end{equation}
    \noindent
where it is assumed that no new individuals enter the group or leave it, or $S(t) + I(t) = N$.

If the individuals are allowed to recover from that disease, then R will be the total number of those that were cured, resulting in the following set of equations:
\begin{gather}
    \begin{aligned}
        \dot{S} &=& -\frac{\beta I S}{N}\\
        \dot{I} &=& \frac{\beta I S}{N} - \gamma I.\\
        \dot{R} &=& \gamma I
    \end{aligned}
\end{gather}
\noindent

In both cases, the parameters $\beta$ and $\gamma$ are essential to characterize the dynamics, as the basic reproductive number $R_0 = \frac{\beta}{\gamma}$ is reliant on them, and characteristic of the disease's threshold: if each infected individual infects more than one susceptible individual, or $R_0 >0$, then there's the onset of the epidemic. Otherwise, if $R_0 <0$, the disease eventually vanishes from the population.

\begin{figure}[htpb]
        \centering
        \includegraphics[width=0.8\linewidth]{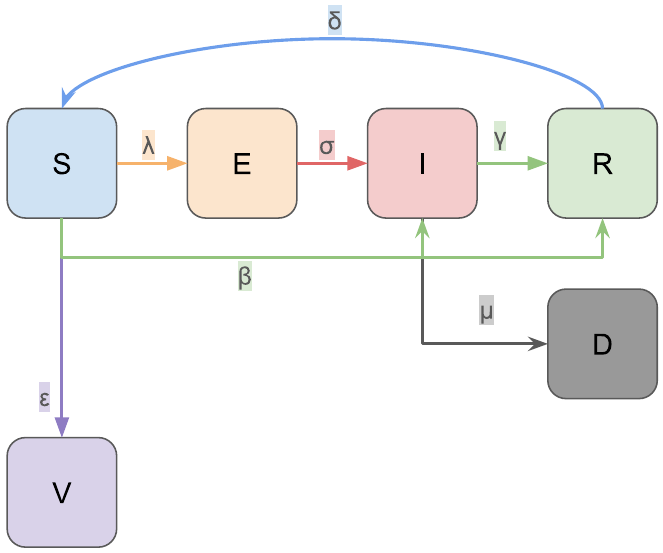}
        \caption{Schematic of the compartmental transition for the selected epidemiological models, which use only parts of all compartments displayed. For their full equations, see \autoref{tab:compartmental models}.}
        \label{fig:compartmental models schematic}
    \end{figure}

Expanding further from the aforementioned SIR and SIS systems, several other compartmental models can be built using different states for the individuals in a network. In this study, the following ones are used, with their equations shown below in \autoref{tab:compartmental models} (see also figure~\ref{fig:compartmental models schematic}).

\begin{itemize}
    \item SIRV \cite{oke2019mathematical}, which vaccinated (V) individuals are also accounted for in the dynamics;
    \item SIRS \cite{hu2019global}, allowing reinfection in the SIR model;
    \item SEIR \cite{annas2020stability}, adding an exposed (E) category to the standard SIR;
    \item SEIRD \cite{korolev2021identification}, expanding the SEIR model by adding a deceased (D) count.
\end{itemize}

Investigating several of these systems allows for broader conclusions on the efficiency of the symbolic regression algorithms examined, as these expanded compartmental models aim to account for the various phases and potential developments of an epidemic, drawing closer to real world disease spreading.

\section{\label{sec:methodology} Methodology}



\subsection{\label{subsec_A_synth_data} Generating synthetic data}

To compare the six chosen symbolic regression methods, data from the dynamic systems described in \autoref{sec:systems for benchmark} were generated. All relevant parameters to simulate these systems and to implement each algorithm are listed in Appendix \ref{AppendixB}. 

It is important to mention that this study does not aim to provide an exhaustive benchmarking of all possible compartmental models in the literature. Moreover, a thorough benchmark would also require an overall study on the effect of parameter selection to the efficacy of SR techniques, and here performance comparisons were restricted to a given set of constants, including epidemiological transition rates for each case (see Appendix \ref{AppendixB} for details).


\subsection{\label{subsec_B_synth_data} Use of in-domain knowledge}



As discussed previously in \autoref{subsec_B_SR_description}, some SR algorithms do have ways to integrate known characteristics of the system under investigation into its regression procedure. When the system is known in its entirety, including its variables and their respective derivatives, the objective of inferring the equation that originated the complete data can be simplified by using in domain knowledge of relationships of such variables. 

In this study, five forms of incorporating prior information during training were considered, each implemented
differently across models when available. \autoref{tab: in domain} below displays the available methods to do so and whether they were used in this study. Those were: (i) whether the variables in the dataset were explicitly selected or split to reflect data that would be descriptive of the known dynamics; (ii) single equation fitting instead of investigating the whole system at once; (iii) selecting the set of operators, which includes mathematical operands and functions; and (iv) choosing the best equation from the list of results of the fitting process, according to what is known of the system, if they were listed as worse according to the model's internal accuracy score. For this last case, only PyKAN required manual adjustments.

\begin{table}[!h]
    \centering
    \caption{\label{tab: in domain} Use of in-domain knowledge of the investigated systems during SR training. A \checkmark indicates that the method is available to the algorithm, and \checkmark\checkmark denotes usage for all systems, unless indicated.}
    \resizebox{0.475\textwidth}{!}{
    \begin{tabular}{c|c|c|c|c}
        \hline
        \diaghead{lalalalalalallalaa}%
        {Method\\}{Algorithm}                & Variable selection    & Single equation fitting & Operator selection    &   Equation fixing                 \\\hline
        GPLearn                              & \checkmark\checkmark  & \checkmark\checkmark\footnote{Not used for the Lorenz, Lotka-Volterra, Pendulum and SIS systems.}  & \checkmark\checkmark   &               \\\hline
        AI Feynman                           & \checkmark\checkmark  & \checkmark\checkmark    & \checkmark\checkmark  &                            \\\hline
        PySINDy                              & \checkmark            &                         & \checkmark\checkmark  &   \checkmark               \\\hline
        PySR                                 & \checkmark            & \checkmark\checkmark    & \checkmark\checkmark  &   \checkmark               \\\hline
        PyKAN                                & \checkmark\checkmark  & \checkmark\checkmark    & \checkmark\checkmark  &   \checkmark\checkmark     \\\hline
        ODEFormer                            & \checkmark            & \checkmark              &                       &                            \\\hline
    \end{tabular}
    }
\end{table}


It is important to mention that, in most common in real-world scenarios, information about the investigated system will likely be absent. Beyond the inability to fine-tune the algorithms by using known properties, this also introduces a significant challenge: during training, calculating derivatives by finite difference methods can introduce (or amplify) noise in data, which in turn complicates the identification of a system. \citet{schaeffer2017sparse} addressed this issue for sparsity-based algorithms through integral formulations that circumvent the need for numerical differentiation, although more studies are needed to further complement other methods.

\subsection{\label{subsec_C_metrics} Metrics and definition of a baseline}

Evaluating the results of a SR method consists in determining whether it is capable of recovering the exact mathematical expression underlying the observed dynamics, that is, verify if the identified expression are be mathematically equivalent to the original equation, along with the estimated parameters approximating the actual values numerically. To this end, this study compared algorithms using three criteria: (i) comparing the structural forms obtained by the symbolic regression algorithms with the original ones; (ii) quantifying the difference of the resulting data on the differential equations to the original system's; and (iii) ascertaining the complexity of the equations obtained by the algorithms.

o quantify structural recovery, a Wilcoxon signed-rank test \cite{Conover1999Jan} was employed to assess whether the SR-generated expressions produced data statistically indistinguishable from that of the true system. Recovery was considered
successful when the null hypothesis of no difference was not rejected at $\alpha = 0.05$ and p-value$>0.05$.

To anchor SR performance to other machine learning methods, a random forest \cite{breiman2001random} algorithm was included as a non-interpretable baseline. For both the original equations and those inferred by SR algorithms, solutions were integrated over the same temporal grid used for data generation, and predictive performance was quantified using the coefficient of determination (R2), following standard practice in the SR literature \cite{LaCava2021Dec}.

Lastly, the complexity of each inferred equation was computed following the discussion in \cite{LaCava2021Dec}, where it is defined as the amount of variables, operators and constants present in the given equation. For multidimensional equations, this metric measures the sum of each dimension's complexity.

\section{\label{sec:results}Results and discussions}

\begin{sidewaystable*}

\vspace{10cm}   

     \centering
       \caption{Original structural form of the dynamic systems, alongside relevant parameters used by the dynamic systems to generate their synthetic data (first column), and equations found by each symbolic regression algorithm (all other columns). Dynamical systems whose structural forms were correctly identified are marked with ``check mark" (\checkmark).}    
       \label{tab:general dynamics results}
    \resizebox{1.0\textwidth}{!}{
    \begin{tabular}{l|l|l|l|l|l|l}
    \hline
    Governing equation & GPLearn &AI-Feynman & PySINDy&PySR &PyKAN  & ODEFormer\\
    \hline

    \begin{tabular}{l} 
        \textbf{Non linear pendulum}\\
        $\dot{\theta}= \omega$ \\
        $\dot{\omega}=-9.8 \sin(\theta) $  \\ 
    \end{tabular}& 
    \begin{tabular}{l}\\{\Large\checkmark} \\  
        $\dot{\theta}= \omega $ \\
        $\dot{\omega}= -9.17sin(1.08\theta)$ \\ 
    \end{tabular}   &
    \begin{tabular}{l}\\ {\Large   \checkmark }  \\  
        $\dot{\theta}= \omega $ \\
        $\dot{\omega}= -9.8\sin(\theta) $ \\ 
    \end{tabular} &
    \begin{tabular}{l}\\{\Large   \checkmark } \\
        $\dot{\theta}= \omega $ \\
        $\dot{\omega}= -9.8\sin(\theta) $ \\ 
    \end{tabular} &
    \begin{tabular}{l}\\{\Large   \checkmark }\\ 
        $\dot{\theta}= \omega$ \\
        $\dot{\omega}=-9.8 \sin(\theta) $ \\ 
    \end{tabular}  &
    \begin{tabular}{l}\\{\Large\checkmark} \\ 
        $\dot{\theta}= \omega $ \\
        $\dot{\omega}=-9.8 \sin(\theta)$ \\ 
    \end{tabular}&
    \begin{tabular}{l} \\ 
        $\dot{\theta}=  1.06 \omega -0.04 \omega (0.05 \omega -13.14 \theta) $ \\
        $\dot{\omega}= -9.41 \theta - \frac{1.2027}{(10.73 -15.52 \theta)} $ \\ 
    \end{tabular}\\
    \hline

    \begin{tabular}{l} \\
        \textbf{Lotka-Volterra}\\
        $\dot{u} = 2 u - 0.5 uv$ \\ 
        $\dot{v} =- v + 0.375uv$\\ 
    \end{tabular}
    & \begin{tabular}{l} \\ {\Large   \checkmark } \\ 
        $\dot{u} = 2u-0.5uv$ \\ 
        $\dot{v} =-v +0.18uv $ \\ 
    \end{tabular}   & 
    \begin{tabular}{l} \\{\Large   \checkmark } \\ 
        $\dot{u} =  2u-0.5uv $ \\ 
        $\dot{v} =-0.99v + 0.33uv $\\ 
    \end{tabular} &
    \begin{tabular}{l} \\ {\Large   \checkmark } \\
        $\dot{u} = 1.94u-0.49uv $ \\ 
        $\dot{v} = -0.95v + 0.37uv$ \\ 
    \end{tabular} & 
    \begin{tabular}{l} \\ {\Large   \checkmark } \\ 
        $\dot{u} = 2.0u -0.5uv $ \\ 
        $\dot{v} = 0.37uv -1.0v$ \\ 
    \end{tabular} & 
    \begin{tabular}{l} \\ 
        $\dot{u} =2u -0.5uv$ \\ 
        $\dot{v} = - v + 0.19uv  $ \\ 
    \end{tabular}&
    \begin{tabular}{l} {\Large   \checkmark } \\ 
        $\dot{u} = 1.5 u -0.4 u v$ \\ 
        $\dot{v} = -1.4 v + 0.2 u v  $ \\ 
    \end{tabular}\\
    \hline
    \begin{tabular}{l} 
        \textbf{Lorenz}\\
        $\dot{x}= 2 (y - x)$ \\
        $\dot{y}=x(1 - z) - y $ \\
        $\dot{z}=x y - 2.6 z $ \\ 
    \end{tabular} & 
    \begin{tabular}{l}\\
        $\dot{x}=  2(y-x)$ \\
        $\dot{y}= x (1-z) - y $ \\
        $\dot{z}=  0.17y - 0.92z $ \\ 
    \end{tabular}   & 
    \begin{tabular}{l}\\
        $\dot{x}= 2(y-x) $ \\
        $\dot{y}= x(1-z)) -y $ \\
        $\dot{z}=0.02+z \sqrt{z}$  \\ 
    \end{tabular}&
    \begin{tabular}{l} \\{\Large\checkmark}\\  
        $\dot{x}= 2 (y - x)$ \\
        $\dot{y}= x(1-z)-y $ \\
        $\dot{z}= xy -2.6z  $\\ 
    \end{tabular} &
    \begin{tabular}{l}\\{\Large\checkmark} \\ 
        $\dot{x}= 2.0(y-x) $ \\
        $\dot{y}= x(1-z)-y$ \\
        $\dot{z}=xy - 0.6z$ \\ 
    \end{tabular} & 
    \begin{tabular}{l}\\{\Large\checkmark} \\ 
        $\dot{x}=2 (y- x)$ \\
        $\dot{y}= x(1 - z) - y$ \\
        $\dot{z}=-xy -2.6z$ \\ 
    \end{tabular}&
    \begin{tabular}{l} \\ 
        $\dot{x}=   1.9 (y-x)$ \\
        $\dot{y}=   -1.2 yz^2$ \\
        $\dot{z}=   0.05y -0.63 z$ \\ 
    \end{tabular}\\
    \hline

    \end{tabular}
}       
    \label{qd:dinami}
\end{sidewaystable*}

\begin{sidewaystable*}  
\vspace{9cm}
    \centering 
     \caption{\label{tab:epi results}Description of the symbolic representations of the approximated epidemic propagation dynamics. Dynamical systems whose structural forms were correctly identified are marked with ``check mark" (\checkmark).}  
    \resizebox{\textwidth}{!}{ 
    \begin{tabular}{l|l|l|l|l|l|l}
    \hline
    Governing equation & GPLearn &AI-Feynman & PySINDy&PySR &PyKAN & ODEFormer\\
    \hline
    
    \multirow{1}{*}{\rotatebox{90}{\textbf{SIS}}}
    \begin{tabular}{l}  
        $\dot{S} = -0.3 S I + 0.1 I $ \\ 
        $\dot{I} = 0.3 S I - 0.1 I$\\ 
    \end{tabular}     &
    \begin{tabular}{l}\\{\Large   \checkmark } \\
        $\dot{S} = -0.29 S I  + 0.1 I  $ \\ 
        $\dot{I} = 0.29 S I  - 0.1 I$\\ 
    \end{tabular}  & 
    \begin{tabular}{l}\\
        $\dot{S} = -0.3 S I + 0.1 I^2 $ \\ 
        $\dot{I} = 0.2 S I - 0.1 I^2$\\ 
    \end{tabular} & 
    \begin{tabular}{l}\\{\Large   \checkmark } \\
        $\dot{S} = -0.3 S I - 0.1 S $ \\ 
        $\dot{I} = 0.3 S I + 0.1 S$\\ 
    \end{tabular}&
    \begin{tabular}{l}\\{\Large   \checkmark } \\ 
        $\dot{S} = -0.3 S I + 0.1 I $ \\ 
        $\dot{I} = 0.3 S I - 0.1 I$\\ 
    \end{tabular} &
    \begin{tabular}{l}\\{\Large   \checkmark }\\ 
        $\dot{S} = -0.3 S I + 0.1I $ \\ 
        $\dot{I} = 0.3 S I - 0.1I$\\ 
    \end{tabular} &
    \begin{tabular}{l}\\ 
        $\dot{S} = 9.01SI (0.03(-1 + 0.14 S)^2 -0.11  S) $ \\ 
        $\dot{I} = 0.09 I (2.1 -3.44 I)$\\ 
    \end{tabular}  \\
    \hline

    \multirow{1}{*}{\rotatebox{90}{\textbf{SIR}}}
    \begin{tabular}{l}
        $\dot{S} = -0.5 S I $ \\ 
        $\dot{I} = 0.5 S I - 0.1 I$\\
        $\dot{R} =  0.1 I$ \\ 
    \end{tabular}  &
    \begin{tabular}{l}\\{\Large   \checkmark } \\ 
        $\dot{S} = -0.5 S I $ \\ 
        $\dot{I} = 0.44 S I - 0.1 I$\\
        $\dot{R} = 0.1 I$ \\ 
    \end{tabular}  & 
    \begin{tabular}{l}\\{\Large    } \\ 
        $\dot{S} = -0.5 S I $ \\ 
        $\dot{I} = SI - I^2 - 0.0007$\\
        $\dot{R} =  0.1 I$ \\ 
    \end{tabular} & 
    \begin{tabular}{l}\\ {\Large   \checkmark } \\ 
        $\dot{S} = -0.5 S I $ \\ 
        $\dot{I} =  0.5 S I - 0.1 I$\\
        $\dot{R} =  0.1I $ \\ 
    \end{tabular} & 
    \begin{tabular}{l}{\Large   \checkmark }\\ 
        $\dot{S} = -0.5 S I $ \\ 
        $\dot{I} = 0.5 S I - 0.1 I$\\
        $\dot{R} =  0.1 I$ \\ 
    \end{tabular}  &
    \begin{tabular}{l}\\{\Large   \checkmark } \\ 
        $\dot{S} = -0.5 S I $ \\ 
        $\dot{I} = 0.5 S I - 0.1 I$\\
        $\dot{R} =  0.1 I$ \\ 
    \end{tabular}   &
    \begin{tabular}{l} \\ 
        $\dot{S} = -0.1 S $ \\ 
        $\dot{I} = 0.5 SI - 0.1R$\\
        $\dot{R} = 0.1I$ \\ 
    \end{tabular}\\
    \hline

    \multirow{1}{*}{\rotatebox{90}{\textbf{SIRV}}}
    \begin{tabular}{l}
        $\dot{S}= -0.5 SI - 0.2 S$ \\
        $\dot{I}= 0.5 SI - 0.1 I$ \\
        $\dot{R}= 0.1 I$ \\
        $\dot{V}= 0.2 S$ \\ 
    \end{tabular}& 
    \begin{tabular}{l} \\ {\Large \checkmark  } \\
        $\dot{S}=  -S I - 0.146 S$ \\
        $\dot{I}= 0.32 S I - 0.09 I$ \\
        $\dot{R}=  0.1 I$ \\
        $\dot{V}= 0.2 S$ \\ 
    \end{tabular}   &
    \begin{tabular}{l} \\ {\Large    } \\ 
        $\dot{S}=  -0.4 S I - 0.2 S \sqrt{I + 1}$ \\
        $\dot{I}= -0.09 I + 0.09 S \sqrt{\sqrt{I}+ 1 }$ \\
        $\dot{R}=  0.1 I$ \\
        $\dot{V}= 0.2 S$ \\ 
    \end{tabular} &
    \begin{tabular}{l} \\  
        $\dot{S}=-0.5 SI -0.2 S $\\
        $\dot{I}= 0.4 S  I -0.2 I V  $ \\
        $\dot{R}= 0  $ \\
        $\dot{V}= 0.2 S  $ \\ 
    \end{tabular} &
    \begin{tabular}{l}\\ {\Large   \checkmark } \\ 
        $\dot{S}= -0.5 SI -0.2 S$ \\
        $\dot{I}= 0.5 SI - 0.1 I$ \\
        $\dot{R}= 0.1 I$ \\
        $\dot{V}= 0.2 S$ \\ 
    \end{tabular} &
    \begin{tabular}{l} \\ {\Large   \checkmark } \\ 
        $\dot{S}= -0.5 SI - 0.2 S$ \\
        $\dot{I}= 0.5 SI - 0.1I$ \\
        $\dot{R}= 0.1 I$ \\
        $\dot{V}= 0.2 S$ \\ 
    \end{tabular}   &
    \begin{tabular}{l} \\ 
        $\dot{S}= -0.3S$ \\
        $\dot{I}= 0.1S - 0.1I$ \\
        $\dot{R}= 1.0 V -2.7R$ \\
        $\dot{V}= 0.2S$ \\ 
    \end{tabular}\\

    \hline
    \multirow{1}{*}{\rotatebox{90}{\textbf{SIRS}}}
    \begin{tabular}{l}
        $\dot{S} = -0.5 S I + 0.2 R$ \\ 
        $\dot{I} = 0.5 S I - 0.1 I$\\
        $\dot{R} = 0.1 I - 0.2 R$ \\ 
    \end{tabular} &
    \begin{tabular}{l} \\{\Large  \checkmark }\\ 
        $\dot{S} =  -0.47 S I + 0.19 R  $ \\ 
        $\dot{I} =  0.49 S I - 0.1 I $\\
        $\dot{R} = 0.10 I - 0.21 R $  \\ 
    \end{tabular}   & 
    \begin{tabular}{l}\\ 
        $\dot{S} =  -0.02S I (S + I^2 - R) - 0.02 S $ \\ 
        $\dot{I} = 1.08I (0.27 S - 0.9)$\\
        $\dot{R} =  I - 0.19 R$ \\ 
    \end{tabular} &  
    \begin{tabular}{l}\\ 
        $\dot{S} =  -0.5 S I + 0.2 R$ \\ 
        $\dot{I} = 0.1 S I$\\
        $\dot{R} = I  + 0.9R$ \\ 
    \end{tabular} & 
    \begin{tabular}{l} \\ {\Large   \checkmark } \\
        $\dot{S} = -0.5 S I + 0.2 R$ \\ 
        $\dot{I} = 0.5 S I - I$\\
        $\dot{R} = 0.1 I - 0.2 R$ \\ 
    \end{tabular} & 
    \begin{tabular}{l}\\ {\Large   \checkmark } \\
        $\dot{S} = -0.317S I + 0.2R - 0.2$ \\ 
        $\dot{I} = 0.3 S I -  1.0I$\\
        $\dot{R} = 1.0 I - 0.2 R$ \\ 
    \end{tabular}   &
    \begin{tabular}{l}\\ 
        $\dot{S} = -0.2 S^2$ \\ 
        $\dot{I} = 0.3I$\\
        $\dot{R} = -1.7R$ \\ 
    \end{tabular}\\
    
    \hline
    \multirow{1}{*}{\rotatebox{90}{\textbf{SEIR}}}
    \begin{tabular}{l} 
        $\dot{S}= -0.5 SI$ \\
        $\dot{E}= 0.5 SI - 0.5 E$ \\
        $\dot{I}= 0.5 E - 0.1 I$ \\
        $\dot{R}= 0.1 I$ \\ 
    \end{tabular} &
    \begin{tabular}{l} \\{\Large   \checkmark  }\\ 
        $\dot{S}= -0.5 S I$ \\
        $\dot{E}= 0.48 S I - 0.48 E $ \\
        $\dot{I}= 0.45 E - 0.096 I$ \\
        $\dot{R}= 0.096 I$ \\ 
    \end{tabular}     & 
        \begin{tabular}{l} \\ 
        $\dot{S}= -0.5 SI$ \\
        $\dot{E}= 0.5 SI - 0.5 E$ \\
        $\dot{I}= 0.27 \left(E + \frac{E^2}{I+E}\right)$ \\
        $\dot{R}= \arcsin\!\left(0.09\, I\right)$ \\ 
    \end{tabular}   &
    \begin{tabular}{l} \\ {\Large  \checkmark  }\\ 
        $\dot{S}=   -0.5 S I$ \\
        $\dot{E}=   0.5 S I - 0.5 E$ \\
        $\dot{I}=   0.5 E -0.1 I$\\
        $\dot{R}=   0.1 I$ \\ 
    \end{tabular} & 
    \begin{tabular}{l} \\ {\Large   \checkmark } \\ 
        $\dot{S}= -0.5 SI$ \\
        $\dot{E}= 0.5 SI - 0.5 E$ \\
        $\dot{I}= 0.5 E - 0.1 I$ \\
        $\dot{R}= 0.1 I$ \\ 
    \end{tabular}&
    \begin{tabular}{l} \\{\Large   \checkmark } \\ 
        $\dot{S}= -0.5IS $ \\
        $\dot{E}= -0.5E + 0.5 IS$ \\
        $\dot{I}= 0.5E - 0.1I$ \\
        $\dot{R}= 0.1I$ \\ 
    \end{tabular}   &
    \begin{tabular}{l} \\
        $\dot{S}= -1.3 SE$ \\
        $\dot{E}= 0.1E - 0.7ER$ \\
        $\dot{I}= 0.4E - 0.1I$ \\
        $\dot{R}= 1.5I - 28.1SIR$ \\ 
    \end{tabular}\\
    
    \hline
    \multirow{1}{*}{\rotatebox{90}{\textbf{SEIRD}}}  
    \begin{tabular}{l} 
        $\dot{S}= -0.5 SI$ \\
        $\dot{E}= 0.5 SI - 0.2 E$ \\
        $\dot{I}= 0.2 E - 0.2I$ \\
        $\dot{R}= 0.1 I$ \\
        $\dot{D}= 0.1 I$ \\ 
    \end{tabular}    &  
    \begin{tabular}{l} \\
        $\dot{S}= -0.5 SI$ \\
        $\dot{E}=  0.1SE - 0.06I$ \\
        $\dot{I}= (E + 0.06)*(E - I)$ \\
        $\dot{R}= 1.0 I$ \\
        $\dot{D}= 0.1 I$ \\ 
    \end{tabular}   & 
    \begin{tabular}{l} \\ 
        $\dot{S}= -0.3 SI$ \\
        $\dot{E}= -0.01(E*((E/I)+1))  $ \\
        $\dot{I}= 0.2 * E - 0.2 I $\\
        $\dot{R}= 0.1 I $ \\
        $\dot{D}= 0 $ \\ 
    \end{tabular}&
    \begin{tabular}{l} \\ {\Large  \checkmark  }\\  
        $\dot{S}= -0.5S I$ \\
        $\dot{E}=  0.5 SI -0.2 E$ \\
        $\dot{I}=  0.2 E - 0.2 I$ \\
        $\dot{R}= 0.1 I$ \\
        $\dot{D}= 0.1 I$ \\ 
    \end{tabular} &
    \begin{tabular}{l} \\ {\Large   \checkmark } \\ 
        $\dot{S}= -0.5 SI$ \\
        $\dot{E}= 0.5 SI - 0.2 E$ \\
        $\dot{I}= 0.2 E - 0.2I$ \\
        $\dot{R}= 0.1 I$ \\
        $\dot{D}= 0.1 I$ \\ 
    \end{tabular} & 
    \begin{tabular}{l}\\ {\Large   \checkmark } \\  
        $\dot{S}= -0.5IS$ \\
        $\dot{E}= -0.2E + 0.5IS$ \\
        $\dot{I}= 0.2 E - 0.2I$ \\
        $\dot{R}= 0.1 I$ \\
        $\dot{D}= 0.1 I$ \\ 
    \end{tabular}&
    \begin{tabular}{l}\\  
        $\dot{S}= -0.4 S I$ \\
        $\dot{E}= 1.0I - 0.9E$ \\
        $\dot{I}= 0.1I - 0.8I(-0.2 - 1.0D)^2$ \\
        $\dot{R}= 0.1 E $ \\
        $\dot{D}= 0.1I$ \\ 
    \end{tabular}\\

    \hline
    \end{tabular}
}
\end{sidewaystable*}

\begin{figure*}[!t]
        \includegraphics[width=0.7\textwidth]{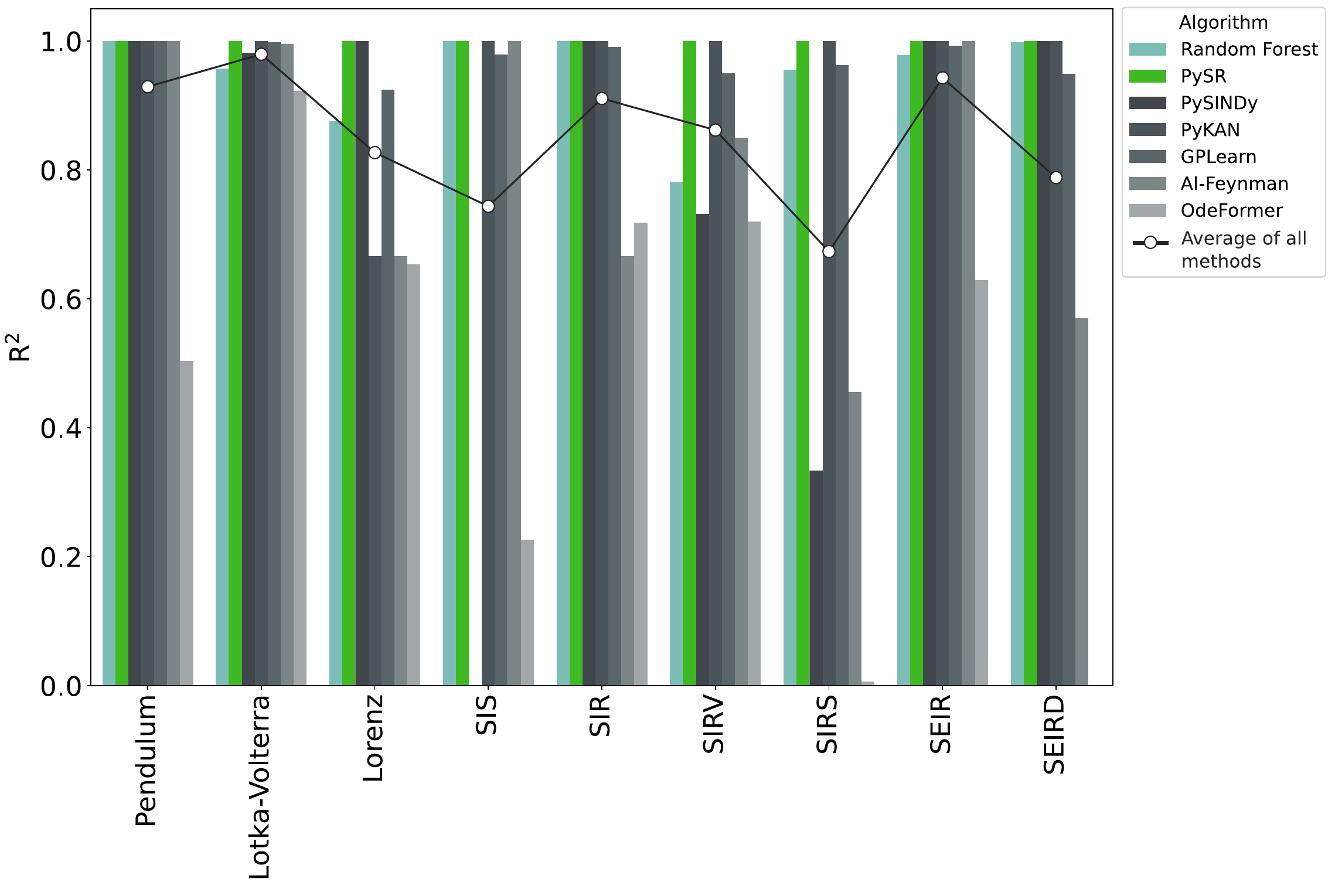}
        \caption{\label{fig:r2} Performance of the symbolic regression models in terms of $R^2$, with a black line displaying an average over all methods for a given system.}
\end{figure*}

Following the specifications of \autoref{sec:methodology}, the results in Table \ref{tab:general dynamics results} and \ref{tab:epi results} were obtained. For all the tables shown below, dynamic systems whose structural forms were correctly identified, when compared to the original equations, are marked with a ``check mark" (\checkmark), while approximations or incorrect descriptions to these systems are left without these markings. 

The results show that all employed symbolic regression models, except for ODEFormer and AI Feynman, demonstrated sufficient capability to identify the symbolic representations of the majority of dynamical systems investigated, specially when considering their effectiveness in the description of structural forms from epidemiological dynamics. 

\begin{table}[!t]
	\centering 
	\caption{\label{tab:overall results all models} Summary of the main results of \autoref{tab:general dynamics results} and \autoref{tab:epi results}. A checkmark indicates that the structural form of the system was successfully identified, while double checkmarks list a result that showed no statistically significant differences compared to the original dynamics, according to the Wilcoxon test. Cells with added numbers show and increase (or decrease) in complexity when compared to the original system.}
	
	\resizebox{0.495\textwidth}{!}{\begin{tabular}{c | c | c |c |c |c | c| c}
			\hline
			\multicolumn{2}{c|}{\textbf{System}} & \multicolumn{6}{c}{\textbf{Symbolic regression method}} \\
			\multicolumn{1}{c}{Name} & \multicolumn{1}{c|}{Complexity} & \multicolumn{1}{c}{GPLearn} & \multicolumn{1}{c}{AI-Feynman} & \multicolumn{1}{c}{PySINDy}& \multicolumn{1}{c}{PySR}& \multicolumn{1}{c}{PyKAN}& \multicolumn{1}{c}{ODEFormer}\\
			\hline        
			Non-linear pendulum    &    8       & \checkmark  \checkmark    & \checkmark \checkmark    & \checkmark  \checkmark   & \checkmark  \checkmark     &  \checkmark \checkmark   &  +13        \\
			\hline        
			Lotka-Volterra         &    19      & \checkmark                         &  \checkmark                       & \checkmark                        & \checkmark \checkmark    &  \checkmark                       &  \checkmark   \\
			\hline
			Lorenz                 &    22      &  \checkmark  \checkmark   &            -2                     & \checkmark                        & \checkmark \checkmark    &  \checkmark                       &    -3         \\
			\hline
			SIS                    &    20      & \checkmark     &  +4       & \checkmark \checkmark                       &  \checkmark \checkmark   &\checkmark  \checkmark    &  +3           \\
			\hline
			SIR                    &    21      &  \checkmark    &  +2  &  \checkmark                       & \checkmark \checkmark    & \checkmark  \checkmark   &  +2           \\
			\hline
			SIRV                   &    27      &  \checkmark    &  +8   &\checkmark  \checkmark  & \checkmark \checkmark    & \checkmark  \checkmark   &  +5           \\
			\hline
			SIRS                   &    28      &     \checkmark       &     -2 &\checkmark  \checkmark & \checkmark &  \checkmark \checkmark &  +5           \\
			\hline
			SEIR                   &    35      &   \checkmark       & -4    &\checkmark  \checkmark & \checkmark  \checkmark   & \checkmark \checkmark  &  -15          \\
			\hline
			SEIRD                  &    39      & - 10   &    -11   & \checkmark  \checkmark &  \checkmark  \checkmark  & \checkmark \checkmark                        &  -6           \\
			\hline
		\end{tabular}
	}
\end{table}
PySR was the best performing algorithm in all aspects, successfully identifying the correct structural form of all systems, and all but one with significant differences in the resulting parameters, making this algorithm the most versatile and accurate of all those tested.

PySINDy and PyKAN also identified all systems, but not as accurately when considering the results of the Wilcoxon test, with PyKAN obtaining SIR's system of equations without significant differences. Although these methods produce slightly inferior results according to \autoref{fig:r2}, they prove to be sufficiently accurate in describing the system's equations of motion. However, it is worth noting that they were more sensitive to changes in the parameter used during the search: while in KAN stronger regularization favors the recovery of the approximate structural form, in some cases it seems to hinder parameter optimization. In PySINDy, on the other hand, optimizing regularization parameters may be the solution to this problem.

GPLearn only could not recover one set of equations (SEIRD), and although most of its results proved to be significantly different from the original dynamics according to the Wilcoxon test, it is still capable of providing the correct structural form of other systems. However, without proper parameter selection, its output equations tend to be more complex than the original structural forms, featuring excessive multiplication and combination of terms, which can be a result of the way GP methods increase the complexity of equations (see Sec. \ref{sec:GPLearn}).  



AI Feynman recovers most equations from the Lorenz, Lotka-Volterra and non-linear pendulum systems, only missing the correct structural form on the z-axis of the former. However, the performance for the compartmental epidemiological systems was subpar, with none of the obtained results reflecting the original dynamics. Inspecting them further, it can be seen that, for SIS, SIR and SIRV, AI Feynman outputs equations of higher complexity than the originating expressions, with the pattern reversed for SIRS, SEIR and SEIRD. This instability in solutions points to the model not being as effective for this particular type of data.

Lastly, ODEFormer only identify one of the systems included in this study, while also having below-average $R^2$ in most cases. This algorithm also inferred equations with degrees of complexity that in some cases varied significantly when compared to the original equations, with a notable decrease in performance for systems with higher dimensions. 

Overall, although some of the resulting equations from the symbolic regression algorithms were only approximations to the real ones, the comparison to the originating systems revealed that these differences were generally small, which can be verified when integrating the resulting equations in the same time frame as the original data and calculating $R^2$ to infer regression efficiency. The results shown in \autoref{fig:r2} indicates that most algorithms can at least effectively capture the overall dynamics and provide an overview at the governing equations for varying phenomena.

\subsection{\label{noise comparisons} The effects of noise in symbolic regression}    

Comparing the chosen algorithms to a set of dynamical systems also requires understanding the effects of corrupted data to the performance of these methods. To do so, two of such systems, the Lotka-Volterra model for predator-prey dynamics and the SIR compartmental model, were chosen to compare the effects of adding noise to data and the resulting effectiveness of the regression performed by all algorithms. The choice for these two systems was based on the results of \autoref{tab:overall results all models}, where almost all algorithms effectively obtained the correct equations for both systems. Beyond adjusting the datasets to add gaussian distributed noise, no hyperparameters (see Tables \ref{tbl:synthetic} and \ref{tab:parametros_epi}) were altered.

    \begin{figure}[htbp] 
    \centering
        \begin{minipage}{.4\textwidth}
            \centering
            \includegraphics[width=1\linewidth]{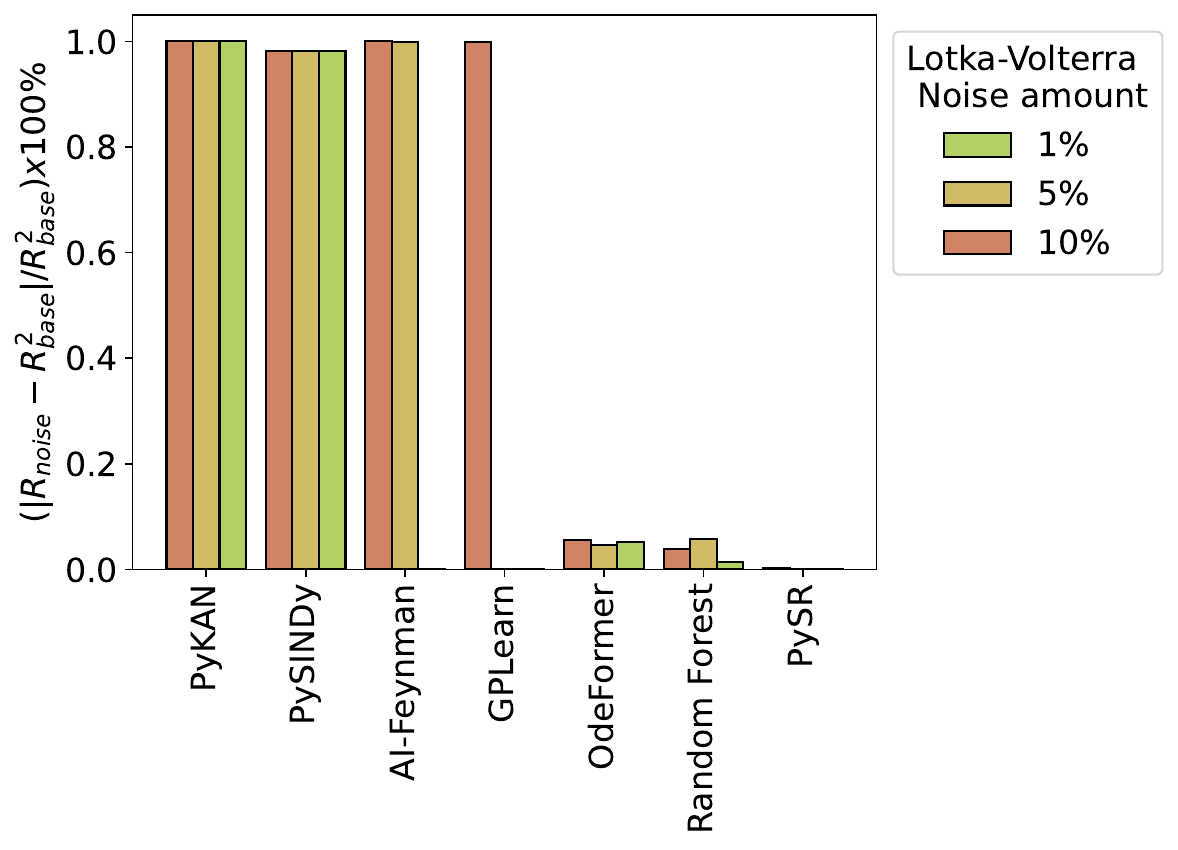} 
        \end{minipage}
        \begin{minipage}{.4\textwidth}
            \centering
            \includegraphics[width=1\linewidth]{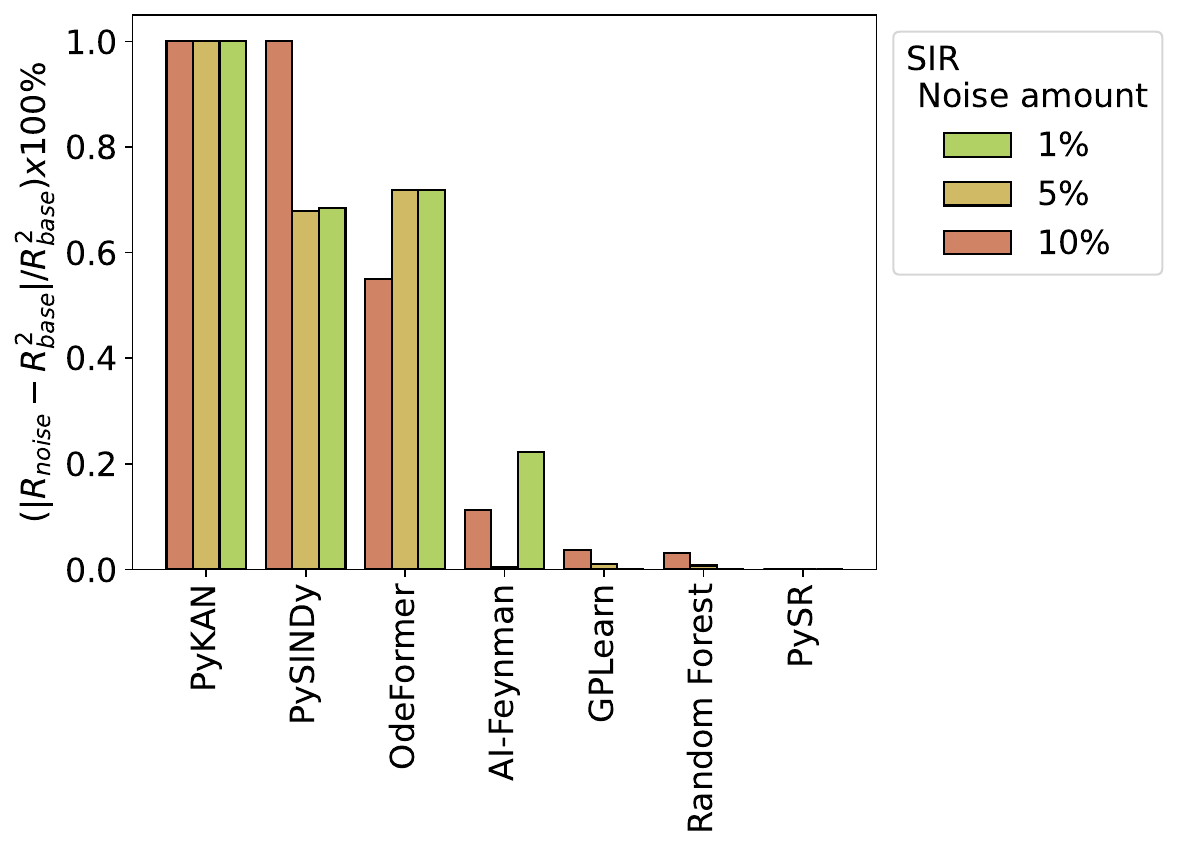} 
        \end{minipage}
       \caption{\label{fig:noise} Resulting differences in $R^2$ when adding gaussian noise to the synthetic data of a (top) Lotka-Volterra and (bottom) SIR system. }     
    \end{figure}

Notably, PyKAN was the most affected by noisy data, presenting a complete loss in $R^2$ for both systems using any amount of noise, pointing to a rather poor generalization of results by this method. In contrast, PySR was the least affected by noise in data, with almost negligible differences in $R^2$ for all cases.

AI-Feynman, PySINDy and GPLearn also presented spikes of performance loss, although not as severe as in PyKAN's case and mostly contained in the Lotka-Volterra system. 

ODEFormer had  below-average $R^2$ loss, with an exception for SIR with 1\% added noise: in it, the model had $R^2=0$ for two components of that system, which presented a significant increase in complexity (37 against 21) compared to the baseline SIR system.


It is important to mention that the results of \autoref{fig:noise} are not equivalent to a systematic benchmark of these techniques: the aim of this particular study was to only increase irregularities on the original data and verify differences in performance. To aim for a thorough benchmark, it would be essential to consider the extensions and settings that can be used to attenuate the effect of noise during their execution, which are present in most SR methods investigated. Below, algorithms that include such strategies to tackle noisy data are discussed:

\begin{itemize}
    \item The original authors of ODEFormer \cite{DAscoli2023Oct} do thorough comparisons by adding different amounts of noise to several systems, showing that the algorithm is robust to noise up to $\sigma=0.05$ (or 5\% added noise) over several systems;
    \item PySINDy provide noise mitigation strategies on their differentiation methods \cite{Kaptanoglu2022}, which is the most sensitive to noise stage of this algorithm;
    \item \citet{Raghav2024Jul} recently extended the original GPLearn method to perform more accurately on noise-sensitive tasks, while also enhancing the original strategy to be able to accommodate user interactivity;
    \item PySR has a built-in denoise module that can be applied by setting the flag \texttt{denoise=True} when initializing the model. More details on the implementation of this method is given on \cite{cranmer2023interpretable}.
    
\end{itemize}

\subsection{\label{AppendixB} Computational complexity}  

\begin{figure}[htbp]
    \includegraphics[width=\linewidth]{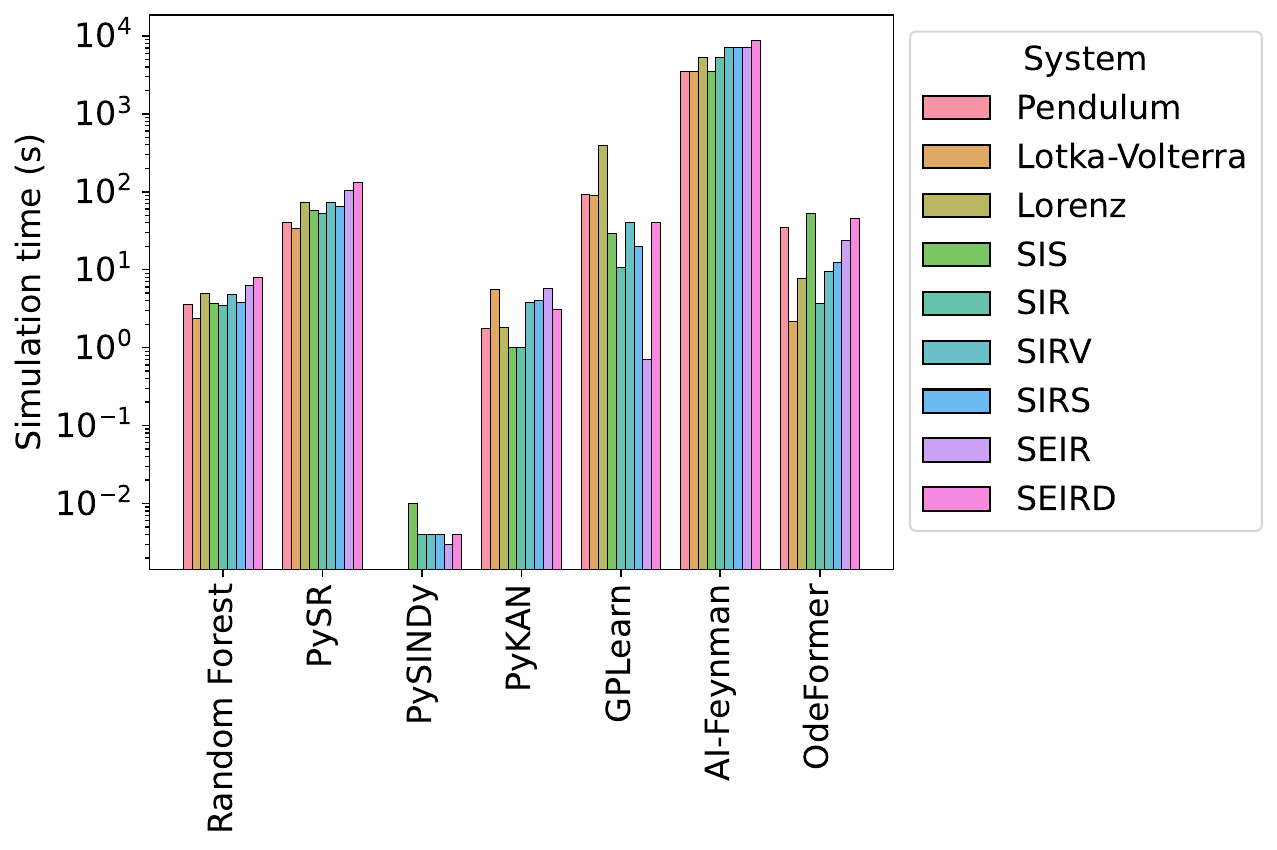}
    \caption{\label{fig:sim times} Processing time for each combination of algorithm and dynamical system, in seconds. For PySINDy, $t=0$s when simulating a Lorenz, Lotka-Volterra and simple pendulum systems.}
\end{figure}

All algorithms are implemented in different ways, and as such the amount of resources, including operation time, may vary. To quantitatively ascertain their computational complexity according to these needs, each individual simulation was timed, with results shown in \autoref{fig:sim times}.

AI-Feynman is, by far, the slowest of the methods analyzed, with each system requiring up to two hours and a half to be executed. The second slowest method is GPLearn, which averaged 75 seconds for each simulation. Contrasting these numbers, PySINDy ran almost instantly for some systems, and kept a very low time average of $5\times10^{-3}$ seconds for others. The rest of the algorithms ran from around a few seconds up to two minutes. These figures are proportionally similar to those listed in the benchmark by \citet{LaCava2021Dec}, although the dataset sizes and computer specifications\footnote{Computer specifications: Fedora Linux 42 (64-bit), 24 × AMD Ryzen 9 9900X 12-Core Processor, 64GB RAM @3600MHz, NVIDIA GeForce RTX 2060 SUPER, B850M D3HP Motherboard.} from both studies are different.






\section{\label{sec_VII_conclusions} Conclusions and outlook}

This study provided a review of both the historic background of symbolic regression and state-of-the-art algorithms currently used by this emerging technique, while also benchmarking several of them in the task of recovering the governing equations of non-linear dynamics. Among all available symbolic regression algorithms, GPLearn, AI-Feynman, PySINDy, PySR, PyKAN  and ODEFormer were selected for these detailed comparisons, using synthetic data of physical, biological and epidemiological systems, with the latter being an application still underexplored for these models. The methodology employed, along with the results obtained through it, allowed the identification of the best overall symbolic regression algorithms, while also presenting an efficient application of such techniques in a new domain.

PySR proved to be the most robust algorithm, recovering the correct structural form in all tested systems and also featuring as the best performing across all metrics: its ability to consistently reconstruct the underlying structure from the data, combined with its computational efficiency, consolidated its superior performance. Only in one case the resulting expressions show statistically significant differences from the original dynamics, although these discrepancies were numerically insignificant.

PySINDy and PyKAN, algorithms that allow for greater interactivity on their training process, also performed well, correctly identifying most dynamic systems. However, in a slightly larger number of cases, the results generated by these methods exhibited significant differences compared to the original equations. It must also be noted that both, and specially PyKAN, had a significant drop in performance when corruption in the form of noise was added to the original data, pointing to a lesser capability for generalization when compared to the top performing method.

Other methods had varying degrees of success, with use cases that can surpass the top performing approaches or supplement them. However, ODEFormer and AI Feynman were two notable outliers when compared to all others. The former, being the first multivariate transformer-based methodology available, thus allowing for inference flexibility that can be further improved upon, has as its current role limited to that of a hypothesis generator, which corroborates \cite{DAscoli2023Oct}. Moreover, although this transfer learning method avoids noise amplification problems from differentiation methods (see \autoref{subsec_B_synth_data}, it may also represent a comparative disadvantage in comparison to algorithms that incorporate a slice of the system's directly in their training stages \cite{DAscoli2023Oct}.

As for AI Feynman, it should be noted that it is the most resource intensive considering the amount of time required to perform simulations. They are also the oldest implemented algorithms from this selection and, in AI-Feynman's case, it is important to mention that the source code, made available by the authors of the methodology themselves in \cite{AI-Feynman-implementation}, has not been maintained in a few years, with the last update dating from 2020. Future users of the code must be aware that technical difficulties may arise from the algorithm relying on dependencies that are out of date.

The results of this study highlights the potential of symbolic regression, suggesting its potential as a new methodology for the modeling of data related to epidemic dynamics, and likely being flexible enough for other domains of knowledge, as current literature indicates. Moreover, updates to these algorithms or the development of new ones could bring even better results when comparing to the already top performing methods, overcoming current limitations and enabling more accurate dynamic recovery and better forecasting results in the future. The continuation of this study is essential to assess the capacity of these methods in describing dynamic systems and, consequently, their contribution to the description of epidemic spreading in real populations.

As for the current limitations of such approach, it is important to mention that even with the current advances to this machine learning field, recovering equations from data through symbolic regression cannot be done in polynomial time, characterizing it as a NP-hard problem \cite{virgolin2022symbolic}. This should be taken into consideration when adding large sets of basis functions and their several permutations to these methods. 

Moreover, this study did not exhaustively benchmarked all methods included, and given the data driven nature of the approach, expanding the comparisons to a larger array of parameters and initial conditions of the chosen systems, and also adding others, would be necessary to further discuss whether a certain method is superior to another, and in which cases. This is specially true when dealing with systems with known chaotic behavior, which includes the investigated Lorenz attractor. Future work can expand on this topic, possibly as an interactive benchmark platform analogue, or as a complement to, SRBench \cite{LaCava2021Dec}.

Finally, although the current results point to a largely successful use of these methods for several systems, the performance for real world datasets can suffer from both limited knowledge of the system's true behavior and the possible imprecision of numerical differentiation methods. Multicollinearity between variables must also be assessed through the appropriate statistical methods, such as Variance Inflation Factors (VIF) and others \cite{Chan2022Apr, Castillo2005Jun}, before applying these algorithms to unknown datasets. In conclusion, future research employing this method for equation discovery should consider the most robust and up to date techniques to tackle these issues, alongside a thorough investigation of the identifiability of the system in question before providing concluding evidence as to the usefulness of symbolic regression to real world data. 


\section*{Data availability}
    All relevant data to this review can be found in \cite{Repo_code}.

\begin{acknowledgments}

Beatriz Brum thanks CAPES for the financial support provided (process number 33001014). Luiza Lober thanks São Paulo Research Foundation (FAPESP) through grants 2022/16065-3 and 2013/07375-0. Francisco A. Rodrigues acknowledges CNPq (grant 308162/2023-4) and FAPESP (grant 20/09835-1) for the financial support given for this research. This study was financed in part by the Coordenação de Aperfeiçoamento de Pessoal de Nível Superior – Brasil (CAPES) – Finance Code 001.


\end{acknowledgments}

\appendix

\newpage

\section{\label{AppendixB} Parameter tables}        

Tables \ref{tbl:synthetic} describe the parameters, initial conditions and time intervals used to generate synthetic data. The \textsc{solve\_ivp} function from the \texttt{SciPy} library was used to numerically integrate these equations over time. The selection of these parameters was made in a case-by-case basis, with the intent of representing a realistic scenario for each system. 

As for the SR algorithm's settings, \autoref{tab:parametros classicas} and \autoref{tab:parametros_epi} list the employed parameters for the non-linear pendulum, Lotka-Volterra and Lorenz systems, and the epidemiological models, in that order. Any other available parameters that were not mentioned were kept as default. These settings were chosen to allow for the best performance possible, and kept the same when investigating the impact of noise in these algorithms (see \autoref{noise comparisons}). 

\begin{table}[!h]
        \centering
        \caption{\label{tbl:synthetic} Parameters employed for solving the Lorenz attractor, non-linear pendulum and predator-prey (Lotka-Volterra) dynamics.}        
        \resizebox{0.38\textwidth}{!}{
        \begin{tabular}{c|c|c|c}
            \hline
                Parameters          & Nonlinear pendulum        & Lotka-Volterra   &   Lorenz\\  \hline
                
                \multirow{3}{*}{\textbf{Initial conditions}}
                &   $\omega =0$     &   $u=20$  &   $x_0=0.6$  \\
                &   $\theta = 45$   &   $v=5$   &   $y_0=2.0$  \\
                &                   &           &   $z_0=1.0$  \\
                \hline
                \multirow{4}{*}{\textbf{Coefficients}}
                &   $\text{g} =9.8$     &   $\alpha=2.0$    &   $\sigma=2.0$    \\
                &   $\text{l} = 1.0$    &   $\beta=0.5$     &   $\rho=1.0$      \\
                &                       &   $\gamma=1.0$    &   $\beta=2.6$     \\
                &                       &   $\delta=0.375$  &   \\
                \hline
                
                \textbf{Simulation window}    & [0,5] & [0,7.5]  & [0,5]\\
                \hline
                \textbf{Time step size}      & 2E-3  & 1E-1     &    2E-3\\\hline  
        \end{tabular}
        }
    \end{table}

\begin{table}[!h]
        \caption{\label{tbl:epidemic} Parameters employed to generate data on the chosen compartmental epidemic models (\ref{sec:epi description}). Compartment sizes are listed as fractions of the total number of individuals.}
        \resizebox{0.42\textwidth}{!}{
        \begin{tabular}{c|c|c|c|c|c|c}
                \hline
                Parameters & SIS & SIR & SIRV  & SIRS & SEIR & SEIRD \\  \hline
                
                \multirow{5}{*}{\textbf{Initial compartment size}}
                &$S_0 = 0.99$   & $S_0 = 0.99$  &   $S_0 = 0.94$    &  $S_0 = 0.99$   & $S_0 = 0.8$  & $S_0 = 0.99$  \\
                &$I_0 = 0.01$   & $I_0 = 0.01$  &   $I_0 = 0.01$    &  $I_0 = 0.01$   & $E_0 = 0.1$  & $E_0 = 0$     \\
                &               & $R_0 = 0$     &   $R_0 = 0$       &  $R_0 = 0$      & $I_0 = 0.1$  & $I_0 = 0.01$  \\
                &               &               &   $V_0 = 0.05 $   &                 & $R_0 = 0$    & $R_0 = 0$     \\
                &               &               &                   &                 &              & $D_0 = 0$     \\
                \hline
                \multirow{5}{*}{\textbf{Coefficients}}
                &$\beta=0.3$  &  $\beta=0.5$  &  $\beta=0.5$    & $\beta=0.5$   & $\beta=0.5$   &$\beta=0.5$   \\
                &$\gamma=0.1$ &  $\gamma=0.1$ &  $\gamma=0.1$   & $\gamma=0.1$  & $\sigma=0.5$  &$\sigma=0.2$   \\
                &             &               &  $\epsilon=0.5$ & $\delta=0.2$  & $\gamma=0.1$  &$\gamma=0.1$   \\
                &             &               &                 &               &               &$\mu=0.1$   \\
                \hline
                \textbf{Simulation window} & [0, 100] &  [0,75] & [0, 35] &   [0,60] &   [0,80]    &[0,120]\\
                \hline 
                \textbf{Time step size}    & 1E-1     & 1E-1    & 1E-1    & 1E-1     & 1E-1        &  1E-1 \\
                \hline
        \end{tabular}
        }
\end{table}

\begin{table*}[htbp]
    \centering 
    \caption{\label{tab:parametros classicas} Settings and parameters used when running the selected SR algorithms for the Lorenz attractor, non-linear pendulum and predator-prey (Lotka-Volterra) dynamics.}
    \label{Wilcoxon}

    \resizebox{\textwidth}{!}{
        \begin{tabular}{lcccccc}
            \toprule[1.2pt] 
            SR model & Parameters & \multicolumn{3}{c}{\textbf{Dynamic System}} \\
            \cline{3-6}
            & & Lorenz & Non-linear pendulum & Lotka-Volterra & &\\
            \hline
            \multirow{7}{*}{\rotatebox{90}{\textbf{GPLearn}}} & population\_size & 5000 & 5000 & 5000 & & \\
            & generations & 50 & 100 & 50 & & \\
            & tournament\_size & 50 & 100 & 50 & & \\
            & stopping\_criteria & 0.01 & 0.01 & 0.01 & & \\
            & p\_crossover & 0.7 & 0.7 & 0.6 & & \\
            & p\_subtree\_mutation & 0.2 & 0.2 & 0.2 & & \\
            & p\_hoist\_mutation & 0.01 & 0.01 & 0.01 & & \\
            & p\_point\_mutation & 0.09 & 0.09 & 0.09 & & \\
            & init\_depth & $(2, 6)$ & $(2, 2)$ & $(8, 9)$ & & \\
            & parsimony\_coefficient & 0.001 & 0.009 & 0.001 & & \\
            & function\_sett & -- & $[\times, +, -, \div, \sin]$ & $[\times, +, -]$& &\\
            & random\_state & 0 & 0 & 0 & & \\
            \cline{2-7}
    	    \multirow{8}{*}{\rotatebox{90}{\textbf{PySINDy}}}	
    	    & library\_functions &  All but $f(x)=x^n$ & \begin{tabular}{c}Fourier (n\_frequencies = 1)\\Polynomial (degree=1)\end{tabular}  &  \begin{tabular}{c} Polynomial (degree=3) \end{tabular} & &\\
        
    	    & feature\_library &   $x, y, z$ &$\theta, \omega$ &  $u, v$ & & \\
        
    	    &differentiation\_method  &  &  FiniteDifference &  \begin{tabular}{c} FiniteDifference (order=2)\end{tabular}  & & \\
     
        	   &\multirow{1}{*}{\rotatebox{90}{Optimizer}} $\begin{cases}
        			  optmizer \\
        			threshold\\
        			alpha\\
        			normalize\_columns\\
        			thresholder\\
        			\nu \\
        			tol
        		\end{cases}$ & $\begin{cases}
        			STLSQ \\
        			0.2\\
        			1E-4\\
        			False \\
        			-\\
        			-\\
        			-\\
        		\end{cases}$ & $\begin{cases}
        			SR3\\
        			0.4\\
        			- \\
        			-\\
        			'l1'\\
        			-\\
        			-
        		\end{cases}$ & $\begin{cases}
        			SR3 \\
        			0.6\\
        			1E-4 \\
        			True\\
        			-\\
        			1\\
        			1E-6
        		\end{cases}$ & & \\
            \cline{2-7}\\
        
      		\multirow{5}{*}{\rotatebox{90}{\textbf{PySR}}	}
            	&Population & [30, 30, 30]& [1, 30]        & & \\
                
                &N interation & [30,30,30]&[30,30] & [10,50,800] \\
                
                & Binary-Operator &[[-, +, *],[-, +, *],[-, +, *]]&     [[*, +],[+, *]]& [[*, +],[-, *]]  \\
                
                & Unary-Operator & [ [ ],[ ],[ ]] & [ [ ],["sin"]]& [[ ],[ ]] \\
                & Nested\_constraints &no &no &no\\
            \cline{2-7}\\
        
            \multirow{5}{*}{\rotatebox{90}{\textbf{PyKAN}}}	& Network topology &
                        $\begin{cases}
            			x:kanpiler \\
            			y:kanpiler\\
            			z:kanpiler \end{cases}$& $\begin{cases}
            			\theta:[1,1] \\
            			\omega:[1,1] \end{cases}$&$\begin{cases}
            			U:kanpiler \\
            			V:kanpiler \end{cases}$ \\
                & seed          & 0    &12     &0   \\
                & $\lambda$      &1E-25 &1E-3   &1E-35 \\
                & steps           &50    &30     &60 \\
                & $\lambda$\_coef &1E-19 &1e-15  &1E-2 \\  
            \cline{2-7}\\
      		\multirow{6}{*}{\rotatebox{90}{\textbf{ODEFormer}}	}
                \\
            	&Beam size     & 100   &   100   & 100 & \\
                \\
                &Temperature   &  0.1  &   0.1   & 0.1 \\
                \\
                \\
            \cline{2-7}\\
        \end{tabular} 
    }
\end{table*}

\begin{sidewaystable*}
  \vspace{9cm}
    \centering
    \caption{\label{tab:parametros_epi} Settings and parameters employed by the SR algorithms when running the selected SR algorithms for the epidemiological systems. }
    \resizebox{1\textwidth}{!}{\begin{tabular}{lcccccccc}
        \toprule[1.2pt] \toprule[1.2pt]
    	SR model	&	Parameters&   \multicolumn{5}{c}{\textbf{Epidemiological models}} \\
    	\cline{3-8}
    	&  &   SIR   & SIS   & SEIR 	& SEIRD 	& 	SIRV 	& 	SIRS 	&  \\
    	\hline
    	\multirow{13}{*}{\rotatebox{90}{\textbf{GPLearn}}}	\\
        & population\_size         &6.5E2   &6.5E2   &1E3 & [no, 5E2, 1E3, 1E3, 1E3] &  [2E3, 1E3, 1E3, 1E3] & [5E2,8E3,1E4]   \\
    	& generations              &150    &120      &[1E2,2E2,1E2,1E2] &[1E2, 1E2, 1E2, 2E2, 1E2] & [3.5E2, 5E2, 5E1,5E1]  & [1E2,1E2,6E2]  \\
    	& tournament\_size         &150    &120      &[1E2,2E2,1E2,1E2]& [no, 1E2, 1E2, 2E2, 1E2]& [3.5E2, 1.5E2, 5E1,5E1] & [5E1,1E2,5E2]  \\
    	& stopping\_criteria       &[1E-2    &[1E-2  &[1E-3, 1E-3, 1E-3, 1E-2] &[no, 1E2, 1E2, 2E2, 1E2] & [1E-2, 1E-2, 1E-2, 1E-3] & [no,1E3,1E3] \\
    	& p\_crossover             &0.7    &0.6   &0.7                        &0.7 & 0.7 & 0.7   \\
    	&  p\_subtree\_mutation    &0.1    &0.2   &0.1                        &0.1 & 0.1 & 0.1  \\
    	& p\_hoist\_mutation       &0.1    &0.01   &[0.1, 0.1, 0.05, 0.05]    &[0.05,0.05,0.05,0.1,0.1]& 0.1 & 0.1 \\
    	& p\_point\_mutation       &0.1    &0.09   &0.1                       & 0.1& 0.1 & 0.1   \\
     
        & init\_depth              &no    &no   &[(2,4),(3, 5),(2,4),(0,5)]&[(no,(3,4),(2,4),(1,2),(1,2)]&[(2,4),(3,3),(2,4),(2,4)]&[no,(3,3),(3,4)]  \\
     
    	& parsimony\_coefficient   &1E-2    &1E-4   & [1E-6, 1E-5, 1E-3, 1E-2] & [no, 1E-4, 1E-4, 1E-1, 1E-1] & 0.01 &  [no, 1E-13, 1E-16] \\
     
    	& function\_sett           &[$\times, +$]    &[$\times,+, -, \div$] & [ $[\times,-]$, no,$[\times,-]$, $[\times,-]$ ] & [[$\times,-$], [$\times, -$], [$\times, -$], [$\times, -$], [$\times, -$]] & [[$ \times,-$], [$\times,-$], [$\times,-$], [$\times,-$]] & [[$ \times,+$], [$\times,-$], [$ \times,-, \neg $]]\\
     
    	& random\_state           &0    &0   &0 & 0& 0 & 0 \\\\	
    	\cline{2-8}\\
        
    	\multirow{4}{*}{\rotatebox{90}{\textbf{AI-Feynman}}	}
    	& BF\_try\_time        & [60s, 60s, 60s]  & 60s  &[240, 300, 3600, 60] & [3600, 360, 60, 60, 60] & [300,300,3600,60] & [3600,60,600] \\
    	& BF\_ops\_file\_type  & 7ops  & 7ops  &7ops &7ops & 7ops & 7ops   \\
    	& polyfit\_deg         & 2  & 2  &2 & [2,2,2,2,1] &2  &2   \\
    	& NN\_epochs           & [400,400,4000] & 400  &[600,4000,600,600] &[1000,600,4000,600, 600] & [4000,4000,600,1000] &[600,600,600] \\\\
    	\cline{2-8}\\
    	\multirow{8}{*}{\rotatebox{90}{\textbf{PySINDy}}}	

        & Functions         &  $x + y,x * y$  & $x + y,x * y$    & $x, x * y$ & $x, x * y$ & $x, x * y$  & $x, x * y$ \\
        & Optimizer         & STLSQ & Orthogonal Matching Pursuit & STLSQ  & STLSQ  & STLSQ  & STLSQ \\
        & Threshold         & 0.1   & 0.6                         & 0.15   & 0.053  & 0.08   & 0.04 \\
        & $\alpha$          & 1E-2  & --                          & 1E-3   & 1E-3   & 1E-5   & 1E-4 \\
        & n\_nonzero\_coefs & --    &  2                          & --     & --     & --     &  --   \\
        & normalize\_columns& True  &  --                         & True   & True   & True   & False   \\\\
        &Differentiation method & \multicolumn{6}{c}{Finite difference, order 2}   \\
        \cline{2-8}\\
    	\multirow{4}{*}{\rotatebox{90}{\textbf{PySR}}	}
            & Maxsize        & 10 & 10 & 10 & 10 & 10 & 10           \\
            &ninteration     &1E3 & 1E3 & 1E3 & 1E3 & 1E3 & 1E3 \\
        	& Binary-Operator& [+, *, -]   &   [+, *, -]   &   [+, *, -]   &   [+, *, -]   &   [+, *, -] & [+, *, -]\\
        	& Unary-Operator & [[ ],[ ],[ ]] & [ [ ],[ ]]& [[ ],[ ],[ ],[ ]] & [ [ ],[ ],[ ],[ ]] &[ [ ],[ ],[ ],[ ]]  &  \\\\
    	\cline{2-8}\\
        
    	\multirow{5}{*}{\rotatebox{90}{\textbf{PyKAN}}}	
            & Network topology   & [kanpiler, kanpiler, [1,1]] & [kanpiler, kanpiler] &  [kanpiler, kanpiler, kanpiler, [1,1]] & [kanpiler, kanpiler, kanpiler, [1,1], [1,1]] & [kanpiler, kanpiler, [1,1], [1,1]] & [kanpiler, kanpiler, [2,1]]\\
            
        	& seed               &  0                    &   12  &        & 12 &12  &12  \\
        	& $\lambda$          & [1E-15, 1E-15, 1E-30]  &   [1E-45,1E-25]   &[1E-10, 1E-10, 1E-10, 1E-20] &[1E-10, 1E-10, 1E-7, 1E-20, 1E-20]  & [1E-3, 1E-3, 1E-20, 1E-20] & [1E-5, 1E-3, 1E-15] \\
         
        	& steps              & [50, 30,300]           &[60,60]     &[40,40,40,300]  &[40,40,50,300,300] & [500,20,300,300]  &[500,500,50]  \\
         
            & $\lambda$\_coef    & [1E-5,1E-5,1E-10]      & [1E-2,1E-12]&    [1E-03, 1E-03, 1E-03, 1E-10]   &[1E-3, 1E-03, 1E-15, 1E-10, 1E-10] &  [1E-15, 1E-15, 1E-10,1E-10]  & [1E-1,1E-15, 1E-10]\\\\
            \cline{2-8}\\

        \multirow{5}{*}{\rotatebox{90}{\textbf{ODEFormer}}}	
        \\\\
        & Beam size                & 100 & 100 & 100 & 100 & 100 & 100 \\
        & Beam temperature         & 0.1 & 0.9 & 0.1 & 0.1 & 0.1 & 0.1 \\\\\\
        \\
    	\bottomrule[1.2pt] \bottomrule[1.2pt] 
    \end{tabular} }     
\end{sidewaystable*}

\clearpage
\bibliography{bibliography}

@phdthesis{minnebo2011empowering,
  title={Empowering knowledge computing with variable selection-On variable importance and variable selection in regression random forests and symbolic regression},
  author={Minnebo, W and Stijven, S},
  year={2011},
  school={Master’s thesis, University of Antwerp, Antwerp, Belgium}
}

@article{kiyani2023framework,
  title={A Framework Based on Symbolic Regression Coupled with eXtended Physics-Informed Neural Networks for Gray-Box Learning of Equations of Motion from Data},
  author={Kiyani, Elham and Shukla, Khemraj and Karniadakis, George Em and Karttunen, Mikko},
  journal={arXiv preprint arXiv:2305.10706},
  year={2023}
}

@article{chen2019revealing,
  title={Revealing complex ecological dynamics via symbolic regression},
  author={Chen, Yize and Angulo, Marco Tulio and Liu, Yang-Yu},
  journal={BioEssays},
  volume={41},
  number={12},
  pages={1900069},
  year={2019},
  publisher={Wiley Online Library}
}

@article{brunton2016discovering,
  title={Discovering governing equations from data by sparse identification of nonlinear dynamical systems},
  author={Brunton, Steven L and Proctor, Joshua L and Kutz, J Nathan},
  journal={Proceedings of the national academy of sciences},
  volume={113},
  number={15},
  pages={3932--3937},
  year={2016},
  publisher={National Acad Sciences}
}

@article{brunton2016sparse,
  title={Sparse identification of nonlinear dynamics with control (SINDYc)},
  author={Brunton, Steven L and Proctor, Joshua L and Kutz, J Nathan},
  journal={IFAC-PapersOnLine},
  volume={49},
  number={18},
  pages={710--715},
  year={2016},
  publisher={Elsevier}
}

@article{koza1992,
	title={Genetic Programming, On the Programming of Computers by Means of Natural Selection. A Bradford Book},
	author={Koza, John R},
	journal={MIT Press},
	year={1992}
}

@article{camps2023discovering,
  title={Discovering causal relations and equations from data},
  author={Camps-Valls, Gustau and Gerhardus, Andreas and Ninad, Urmi and Varando, Gherardo and Martius, Georg and Balaguer-Ballester, Emili and Vinuesa, Ricardo and Diaz, Emiliano and Zanna, Laure and Runge, Jakob},
  journal={arXiv preprint arXiv:2305.13341},
  year={2023}
}

@article{mangan2016inferring,
  title={Inferring biological networks by sparse identification of nonlinear dynamics},
  author={Mangan, Niall M and Brunton, Steven L and Proctor, Joshua L and Kutz, J Nathan},
  journal={IEEE Transactions on Molecular, Biological and Multi-Scale Communications},
  volume={2},
  number={1},
  pages={52--63},
  year={2016},
  publisher={IEEE}
}

@article{champion2020unified,
  title={A unified sparse optimization framework to learn parsimonious physics-informed models from data},
  author={Champion, Kathleen and Zheng, Peng and Aravkin, Aleksandr Y and Brunton, Steven L and Kutz, J Nathan},
  journal={IEEE Access},
  volume={8},
  pages={169259--169271},
  year={2020},
  publisher={IEEE}
}

@article{de2020pysindy,
 title={Pysindy: a python package for the sparse identification of nonlinear dynamics from data},
  author={De Silva, Brian M and Champion, Kathleen and Quade, Markus and Loiseau, Jean-Christophe and Kutz, J Nathan and Brunton, Steven L},
  journal={arXiv preprint arXiv:2004.08424},
  year={2020}
}

@article{kaptanoglu2021pysindy,
  title={PySINDy: A comprehensive Python package for robust sparse system identification},
  author={Kaptanoglu, Alan A and de Silva, Brian M and Fasel, Urban and Kaheman, Kadierdan and Goldschmidt, Andy J and Callaham, Jared L and Delahunt, Charles B and Nicolaou, Zachary G and Champion, Kathleen and Loiseau, Jean-Christophe and others},
  journal={arXiv preprint arXiv:2111.08481},
  year={2021}
}

@article{schmidt2009distilling,
  title={Distilling free-form natural laws from experimental data},
  author={Schmidt, Michael and Lipson, Hod},
  journal={science},
  volume={324},
  number={5923},
  pages={81--85},
  year={2009},
  publisher={American Association for the Advancement of Science}
}

@misc{dubvcakova2011eureqa,
  title={Eureqa: software review},
  author={Dub{\v{c}}{\'a}kov{\'a}, Ren{\'a}ta},
  year={2011},
  publisher={Springer}
}

@article{schmidt2014eureqa,
  title={Eureqa (version 0.98 beta)},
  author={Schmidt, M and Lipson, H},
  journal={Nutonian Inc., Boston MA},
  year={2014}
}

@book{diveev2021machine,
	title={Machine Learning Control by Symbolic Regression},
	author={Diveev, Askhat and Shmalko, Elizaveta},
	year={2021},
	publisher={Springer}
}

@article{udrescu2020ai,
	title={AI Feynman: A physics-inspired method for symbolic regression},
	author={Udrescu, Silviu-Marian and Tegmark, Max},
	journal={Science Advances},
	volume={6},
	number={16},
	pages={eaay2631},
	year={2020},
	publisher={American Association for the Advancement of Science}
}

@article{miyazaki2023,
  title={Application of the symbolic regression program AI-Feynman to psychology},
  author={Miyazaki, Masato and Ishikawa, Ken-Ichi and Nakashima, Ken'ichiro and Shimizu, Hiroshi and Takahashi, Taiki and Takahashi, Nobuyuki},
  journal={Frontiers in Artificial Intelligence},
  volume={6},
  pages={1039438},
  year={2023},
  publisher={Frontiers}
}

@article{golberg1989genetic,
	title={Genetic algorithms in search, optimization, and machine learning},
	author={Golberg, David E},
	journal={Addion wesley},
	volume={1989},
	number={102},
	pages={36},
	year={1989}
}

@book{sivanandam2008genetic,
	title={Genetic algorithms},
	author={Sivanandam, SN and Deepa, SN},
	booktitle={Introduction to genetic algorithms},
	pages={15--37},
	year={2008},
	publisher={Springer}
}

@misc{darwin1859origin,
	title={The Origin of Species by Means of Natural Selection. Popular impression of the corrected copyright edition},
	author={Darwin, CD},
	year={1910},
	publisher={John Murray, London}
}

@incollection{rothlauf2006representations,
	title={Representations for genetic and evolutionary algorithms},
	author={Rothlauf, Franz},
	booktitle={Representations for Genetic and Evolutionary Algorithms},
	pages={9--32},
	year={2006},
	publisher={Springer}
}

@article{quade2018sparse,
  title={Sparse identification of nonlinear dynamics for rapid model recovery},
  author={Quade, Markus and Abel, Markus and Nathan Kutz, J and Brunton, Steven L},
  journal={Chaos: An Interdisciplinary Journal of Nonlinear Science},
  volume={28},
  number={6},
  pages={063116},
  year={2018},
  publisher={AIP Publishing LLC}
}

@misc{stephens2016genetic,
  title={Genetic Programming in Python, with a scikit-learn inspired API: gplearn},
  author={Stephens, Trevor},
  year={2016}
}

@techreport{gudetti2023,
  title={Data-Driven Modeling of Linear and Nonlinear Dynamic Systems for Noise and Vibration Applications},
  author={Gudetti, Jacinth Philemon and Yazdi, Seyed Jamaleddin Mostafavi and Baqersad, Javad and Peters, Diane and Ghamari, Mohammad},
  year={2023},
  institution={SAE Technical Paper}
}

@article{la2021contemporary,
  title={Contemporary symbolic regression methods and their relative performance},
  author={La Cava, William and Orzechowski, Patryk and Burlacu, Bogdan and de Fran{\c{c}}a, Fabr{\'\i}cio Olivetti and Virgolin, Marco and Jin, Ying and Kommenda, Michael and Moore, Jason H},
  journal={arXiv preprint arXiv:2107.14351},
  year={2021}
}

@inproceedings{abdellaoui2021symbolic,
  title={Symbolic regression for scientific discovery: an application to wind speed forecasting},
  author={Abdellaoui, Ismail Alaoui and Mehrkanoon, Siamak},
  booktitle={2021 IEEE Symposium Series on Computational Intelligence (SSCI)},
  pages={01--08},
  year={2021},
  organization={IEEE}
}

@article{wang2019symbolic,
  title={Symbolic regression in materials science},
  author={Wang, Yiqun and Wagner, Nicholas and Rondinelli, James M},
  journal={MRS Communications},
  volume={9},
  number={3},
  pages={793--805},
  year={2019},
  publisher={Cambridge University Press}
}

@article{luo2023application,
  title={The Application of Symbolic Regression on Identifying Implied Volatility Surface},
  author={Luo, Jiayi and Yu, Cindy Long},
  journal={Mathematics},
  volume={11},
  number={9},
  pages={2108},
  year={2023},
  publisher={MDPI}
}

@article{stanislawska2012modeling,
  title={Modeling global temperature changes with genetic programming},
  author={Stanislawska, Karolina and Krawiec, Krzysztof and Kundzewicz, Zbigniew W},
  journal={Computers \& Mathematics with Applications},
  volume={64},
  number={12},
  pages={3717--3728},
  year={2012},
  publisher={Elsevier}
}

@inproceedings{sahoo2018learning,
  title={Learning equations for extrapolation and control},
  author={Sahoo, Subham and Lampert, Christoph and Martius, Georg},
  booktitle={International Conference on Machine Learning},
  pages={4442--4450},
  year={2018},
  organization={PMLR}
}

@inproceedings{langley1977bacon,
  title={BACON: A Production System That Discovers Empirical Laws.},
  author={Langley, Pat},
  booktitle={IJCAI},
  pages={344},
  year={1977},
  organization={Citeseer}
}

@article{zheng2018unified,
  title={A unified framework for sparse relaxed regularized regression: SR3},
  author={Zheng, Peng and Askham, Travis and Brunton, Steven L and Kutz, J Nathan and Aravkin, Aleksandr Y},
  journal={IEEE Access},
  volume={7},
  pages={1404--1423},
  year={2018},
  publisher={IEEE}
}

@article{rudy2017data,
  title={Data-driven discovery of partial differential equations},
  author={Rudy, Samuel H and Brunton, Steven L and Proctor, Joshua L and Kutz, J Nathan},
  journal={Science advances},
  volume={3},
  number={4},
  pages={e1602614},
  year={2017},
  publisher={American Association for the Advancement of Science}
}

@article{breiman2001random,
  title={Random forests},
  author={Breiman, Leo},
  journal={Machine learning},
  volume={45},
  pages={5--32},
  year={2001},
  publisher={Springer}
}

@article{liu2024kan,
	author = {Liu, Ziming and Wang, Yixuan and Vaidya, Sachin and Ruehle, Fabian and Halverson, James and Solja{\ifmmode\check{c}\else\v{c}\fi}i{\ifmmode\acute{c}\else\'{c}\fi}, Marin and Hou, Thomas Y. and Tegmark, Max},
	title = {{KAN: Kolmogorov-Arnold Networks}},
	journal = {arXiv},
	year = {2024},
	month = apr,
	eprint = {2404.19756},
	doi = {10.48550/arXiv.2404.19756}
}

@article{liu2024kan2,
  title={Kan 2.0: Kolmogorov-arnold networks meet science},
  author={Liu, Ziming and Ma, Pingchuan and Wang, Yixuan and Matusik, Wojciech and Tegmark, Max},
  journal={arXiv preprint arXiv:2408.10205},
  year={2024}
}

@incollection{gustafson2005,
	author = {Gustafson, S. and Burke, E. K. and Krasnogor, N.},
	title = {{On improving genetic programming for symbolic regression}},
	booktitle = {{2005 IEEE Congress on Evolutionary Computation}},
	pages = {02--05},
	isbn = {978-0-7803-9363},
	publisher = {IEEE},
	doi = {10.1109/CEC.2005.1554780},
    year = {2005}
}

@article{annas2020stability,
  title={Stability analysis and numerical simulation of SEIR model for pandemic COVID-19 spread in Indonesia},
  author={Annas, Suwardi and Pratama, Muh Isbar and Rifandi, Muh and Sanusi, Wahidah and Side, Syafruddin},
  journal={Chaos, solitons \& fractals},
  volume={139},
  pages={110072},
  year={2020},
  publisher={Elsevier}
}

@article{korolev2021identification,
  title={Identification and estimation of the SEIRD epidemic model for COVID-19},
  author={Korolev, Ivan},
  journal={Journal of econometrics},
  volume={220},
  number={1},
  pages={63--85},
  year={2021},
  publisher={Elsevier}
}

@article{oke2019mathematical,
  title={Mathematical modeling and stability analysis of a SIRV epidemic model with non-linear force of infection and treatment},
  author={Oke, MO and Ogunmiloro, OM and Akinwumi, CT and Raji, RA},
  journal={Communications in Mathematics and Applications},
  volume={10},
  number={4},
  pages={717},
  year={2019},
  publisher={RGN Publications}
}

@article{hu2019global,
  title={Global dynamics of an SIRS model with demographics and transfer from infectious to susceptible on heterogeneous networks},
  author={Hu, Haijun and Yuan, Xupu and Huang, Lihong and Huang, Chuangxia},
  journal={Math. Biosci. Eng},
  volume={16},
  number={5},
  pages={5729--5749},
  year={2019}
}

@misc{Feynman_lectures_online,
	title = {{The Feynman Lectures on Physics}},
	year = {2024},
	month = oct,
	note = {[Online; accessed 7. Nov. 2024]},
	url = {https://www.feynmanlectures.caltech.edu}
}

@article{cranmer2023interpretable,
  title={Interpretable machine learning for science with PySR and SymbolicRegression. jl},
  author={Cranmer, Miles},
  journal={arXiv preprint arXiv:2305.01582},
  year={2023}
}

@article{wong2022automated,
  title={Automated discovery of interpretable gravitational-wave population models},
  author={Wong, Kaze WK and Cranmer, Miles},
  journal={arXiv preprint arXiv:2207.12409},
  year={2022}
}

@article{
virgolin2022symbolic,
title={Symbolic Regression is {NP}-hard},
author={Marco Virgolin and Solon P Pissis},
journal={Transactions on Machine Learning Research},
issn={2835-8856},
year={2022},
url={https://openreview.net/forum?id=LTiaPxqe2e},
note={}
}

@book{Conover1999Jan,
	author = {Conover, W. J.},
	title = {{Practical Nonparametric Statistics, 3rd Edition}},
	year = {1999},
	month = jan,
	isbn = {978-0-471-16068-7},
	publisher = {Wiley},
	address = {Hoboken, NJ, USA},
	url = {https://www.wiley.com/en-us/Practical+Nonparametric+Statistics%2C+3rd+Edition-p-9780471160687},
    pages = {350}
}

@article{Diz-Pita2021Jul,
	author = {Diz-Pita, {\ifmmode\acute{E}\else\'{E}\fi}rika and Otero-Espinar, M. Victoria},
	title = {{Predator{\textendash}Prey Models: A Review of Some Recent Advances}},
	journal = {Mathematics},
	volume = {9},
	number = {15},
	pages = {1783},
	year = {2021},
	month = jul,
	issn = {2227-7390},
	publisher = {Multidisciplinary Digital Publishing Institute},
	doi = {10.3390/math9151783}
}

@article{Lotka-Volterra-Original-Pub,
 ISSN = {00664162},
 URL = {http://www.jstor.org/stable/2096748},
 author = {Peter J. Wangersky},
 journal = {Annual Review of Ecology and Systematics},
 pages = {189--218},
 publisher = {Annual Reviews},
 title = {Lotka-Volterra Population Models},
 urldate = {2025-05-13},
 volume = {9},
 year = {1978}
}

@article{Lorenz1963Mar,
	author = {Lorenz, Edward N.},
	title = {{Deterministic Nonperiodic Flow}},
	journal = {J. Atmos. Sci.},
	volume = {20},
	number = {2},
	pages = {130--141},
	year = {1963},
	month = mar,
	issn = {0022-4928},
	publisher = {American Meteorological Society},
	doi = {10.1175/1520-0469(1963)020<0130:DNF>2.0.CO;2}
}

@article{Haken1963Jun,
	author = {Haken, H. and Sauermann, H.},
	title = {{Nonlinear interaction of laser modes}},
	journal = {Z. Phys.},
	volume = {173},
	number = {3},
	pages = {261--275},
	year = {1963},
	month = jun,
	publisher = {Springer-Verlag},
	doi = {10.1007/BF01377828}
}

@incollection{Shen2025Jan,
	author = {Shen, Bo-Wen},
	title = {{Attractor Coexistence, Butterfly Effects, and Chaos (ABC): A Review of Lorenz and Generalized Lorenz Models}},
	booktitle = {{16th Chaotic Modeling and Simulation International Conference}},
	journal = {SpringerLink},
	pages = {589--610},
	year = {2025},
	month = jan,
	issn = {2213-8692},
	isbn = {978-3-031-60907-7},
	publisher = {Springer},
	address = {Cham, Switzerland},
	doi = {10.1007/978-3-031-60907-7_42}
}

@misc{AI-Feynman-implementation,
	title = {{AI-Feynman}},
    author = {Udrescu, Silviu-Marian and Tegmark, Max},
	journal = {GitHub},
	year = {2020},
	note = {[Online; accessed 22. Jul. 2025]},
	url = {https://github.com/SJ001/AI-Feynman}
}

@misc{Repo_code,
	title = {{GitHub repository for all code used in this study}},
    author = {Brum, B.R and Lober, L.},
	year = {2025},
	month = jul,
    note = {https://github.com/luizalober/review\_symb\_regression/},
	url = {https://github.com/luizalober/review_symb_regression/}
}

@book{Gleiser,
  title={The dancing universe: From creation myths to the big bang},
  author={Gleiser, Marcelo},
  year={2005},
  publisher={UPNE}
}

@article{DAscoli2023Oct,
	author = {D'Ascoli, St{\ifmmode\acute{e}\else\'{e}\fi}phane and Becker, S{\ifmmode\ddot{o}\else\"{o}\fi}ren and Mathis, Alexander and Schwaller, Philippe and Kilbertus, Niki},
	title = {{ODEFormer: Symbolic Regression of Dynamical Systems with Transformers}},
	journal = {arXiv},
	year = {2023},
	month = oct,
	eprint = {2310.05573},
	doi = {10.48550/arXiv.2310.05573}
}

@incollection{Kamienny2022Nov,
	author = {Kamienny, Pierre-Alexandre and D'Ascoli, St{\ifmmode\acute{e}\else\'{e}\fi}phane and Lample, Guillaume and Charton, Fran{\ifmmode\mbox{\c{c}}\else\c{c}\fi}ois},
	title = {{End-to-end symbolic regression with transformers}},
	booktitle = {{Guide Proceedings}},
	pages = {10269--10281},
	year = {2022},
	month = nov,
	publisher = {Curran Associates Inc.},
	doi = {10.5555/3600270.3601016}
}

@article{Vastl2022May,
	author = {Vastl, Martin and Kulh{\ifmmode\acute{a}\else\'{a}\fi}nek, Jon{\ifmmode\acute{a}\else\'{a}\fi}{\ifmmode\check{s}\else\v{s}\fi} and Kubal{\ifmmode\acute{\imath}\else\'{\i}\fi}k, Ji{\ifmmode\check{r}\else\v{r}\fi}{\ifmmode\acute{\imath}\else\'{\i}\fi} and Derner, Erik and Babu{\ifmmode\check{s}\else\v{s}\fi}ka, Robert},
	title = {{SymFormer: End-to-end symbolic regression using transformer-based architecture}},
	journal = {arXiv},
	year = {2022},
	month = may,
	eprint = {2205.15764},
	doi = {10.48550/arXiv.2205.15764}
}

@article{Vaswani2017Jun,
	author = {Vaswani, Ashish and Shazeer, Noam and Parmar, Niki and Uszkoreit, Jakob and Jones, Llion and Gomez, Aidan N. and Kaiser, Lukasz and Polosukhin, Illia},
	title = {{Attention Is All You Need}},
	journal = {arXiv},
	year = {2017},
	month = jun,
	eprint = {1706.03762},
	doi = {10.48550/arXiv.1706.03762}
}

@incollection{VanGael2008Jul,
	author = {Van Gael, Jurgen and Saatci, Yunus and Teh, Yee Whye and Ghahramani, Zoubin},
	title = {{Beam sampling for the infinite hidden Markov model}},
	booktitle = {{ACM Other conferences}},
	pages = {1088--1095},
	year = {2008},
	month = jul,
	publisher = {Association for Computing Machinery},
	address = {New York, NY, USA},
	doi = {10.1145/1390156.1390293}
}

@article{LaCava2021Dec,
	author = {La Cava, William and Orzechowski, Patryk and Burlacu, Bogdan and de Franca, Fabricio and Virgolin, Marco and Jin, Ying and Kommenda, Michael and Moore, Jason},
	title = {{Contemporary Symbolic Regression Methods and their Relative Performance}},
	journal = {Proceedings of the Neural Information Processing Systems Track on Datasets and Benchmarks},
	volume = {1},
	year = {2021},
	month = dec,
	url = {https://datasets-benchmarks-proceedings.neurips.cc/paper/2021/hash/c0c7c76d30bd3dcaefc96f40275bdc0a-Abstract-round1.html}
}

@article{Kaptanoglu2022,
doi = {10.21105/joss.03994},
url = {https://doi.org/10.21105/joss.03994},
year = {2022},
publisher = {The Open Journal},
volume = {7},
number = {69},
pages = {3994},
author = {Alan A. Kaptanoglu and Brian M. de Silva and Urban Fasel and Kadierdan Kaheman and Andy J. Goldschmidt and Jared Callaham and Charles B. Delahunt and Zachary G. Nicolaou and Kathleen Champion and Jean-Christophe Loiseau and J. Nathan Kutz and Steven L. Brunton},
title = {PySINDy: A comprehensive Python package for robust sparse system identification},
journal = {Journal of Open Source Software}
}

@incollection{Raghav2024Jul,
	author = {Raghav, S. Sanjith and Kumar, S. Tejesh and Balaji, Rishiikesh and Sanjay, M. and Shunmuga, C.},
	title = {{Interactive Symbolic Regression - A Study on Noise Sensitivity and Extrapolation Accuracy}},
	booktitle = {{ACM Conferences}},
	pages = {2076--2082},
	year = {2024},
	month = jul,
	publisher = {Association for Computing Machinery},
	address = {New York, NY, USA},
	doi = {10.1145/3638530.3664130}
}

@article{Valipour2021Jun,
	author = {Valipour, Mojtaba and You, Bowen and Panju, Maysum and Ghodsi, Ali},
	title = {{SymbolicGPT: A Generative Transformer Model for Symbolic Regression}},
	journal = {arXiv},
	year = {2021},
	month = jun,
	eprint = {2106.14131},
	doi = {10.48550/arXiv.2106.14131}
}

@article{Qiu2022Oct,
	author = {Qiu, Xing and Xu, Tao and Soltanalizadeh, Babak and Wu, Hulin},
	title = {{Identifiability analysis of linear ordinary differential equation systems with a single trajectory}},
	journal = {Appl. Math. Comput.},
	volume = {430},
	pages = {127260},
	year = {2022},
	month = oct,
	issn = {0096-3003},
	publisher = {Elsevier},
	doi = {10.1016/j.amc.2022.127260}
}

@incollection{Scholl2023,
	author = {Scholl, Philipp and Bacho, Aras and Boche, Holger and Kutyniok, Gitta},
	title = {{The Uniqueness Problem of Physical Law Learning}},
	booktitle = {{ICASSP 2023 - 2023 IEEE International Conference on Acoustics, Speech and Signal Processing, Proceedings}},
	journal = {Technical University of Munich},
	year = {2023},
	publisher = {Institute of Electrical and Electronics Engineers Inc.},
	doi = {10.1109/ICASSP49357.2023.10095017}
}

@article{schaeffer2017sparse,
  title={Sparse model selection via integral terms},
  author={Schaeffer, Hayden and McCalla, Scott G},
  journal={Physical Review E},
  volume={96},
  number={2},
  pages={023302},
  year={2017},
  publisher={APS}
}

@article{Chan2022Apr,
	author = {Chan, Jireh Yi-Le and Leow, Steven Mun Hong and Bea, Khean Thye and Cheng, Wai Khuen and Phoong, Seuk Wai and Hong, Zeng-Wei and Chen, Yen-Lin},
	title = {{Mitigating the Multicollinearity Problem and Its Machine Learning Approach: A Review}},
	journal = {Mathematics},
	volume = {10},
	number = {8},
	pages = {1283},
	year = {2022},
	month = apr,
	issn = {2227-7390},
	publisher = {Multidisciplinary Digital Publishing Institute},
	doi = {10.3390/math10081283}
}

@incollection{Castillo2005Jun,
	author = {Castillo, Flor A. and Villa, Carlos M.},
	title = {{Symbolic regression in multicollinearity problems}},
	booktitle = {{ACM Conferences}},
	pages = {2207--2208},
	year = {2005},
	month = jun,
	publisher = {Association for Computing Machinery},
	address = {New York, NY, USA},
	doi = {10.1145/1068009.1068377}
}

\end{document}